\documentclass[runningheads]{llncs}
\usepackage[table]{xcolor}
\usepackage{graphicx}
\usepackage{tikz}
\usepackage{comment}
\usepackage{amsmath,amssymb} % define this before the line numbering.
\usepackage{color}
\usepackage{float}
\usepackage{subcaption}
\usepackage{multirow}
\usepackage{multicol}
\usepackage{hyperref}  %hyperref still needs to be put at the end!

\usepackage[accsupp]{axessibility}  % Improves PDF readability for those with disabilities.

\newcommand{\norm}[1]{\left\lVert#1\right\rVert}

\begin{document}
% \linenumbers
\pagestyle{headings}
\mainmatter
\def\ECCVSubNumber{4019}  % Insert your submission number here

\title{Spatially Invariant Unsupervised 3D Object-Centric Learning and Scene Decomposition} % Replace with your title

\begin{comment}
\titlerunning{ECCV-22 submission ID \ECCVSubNumber} 
\authorrunning{ECCV-22 submission ID \ECCVSubNumber} 
\author{Anonymous ECCV submission}
\institute{Paper ID \ECCVSubNumber}
\end{comment}
%******************

% \begin{comment}
  \titlerunning{SPAIR3D for Object-Centric Learning and Scene Decomposition}
  \author{Tianyu Wang\inst{1} \and
  Miaomiao Liu\inst{1} \and
  Kee Siong Ng\inst{1}}

  % \author{Tianyu Wang\inst{1}\orcidID{0000-0001-9032-8488} \and
  % Miaomiao Liu\inst{1}\orcidID{0000-0001-6485-3510} \and
  % Kee Siong Ng\inst{1}\orcidID{0000-0003-0701-8783}}

  %
  \authorrunning{TY. Wang et al.}
  \institute{
    Australian National University, Canberra ACT 2601, AU \\ \email{\{Tianyu.Wang2,miaomiao.liu,KeeSiong.ng\}@anu.edu.au} 
  }
  % \end{comment}
%******************
\maketitle

\begin{abstract}
We tackle the problem of object-centric learning on point clouds, which is crucial for high-level relational reasoning and scalable machine intelligence. 
In particular, we introduce a framework,~{\bf SPAIR3D}, to factorize a 3D point cloud into a spatial mixture model where each component corresponds to one object. 
To model the spatial mixture model on point clouds, we derive the~\emph{Chamfer Mixture Loss}, which fits naturally into our variational training pipeline. 
Moreover, we adopt an object-specification scheme that describes each object’s location relative to its local voxel grid cell. 
Such a scheme allows~{\bf SPAIR3D} to model scenes with an arbitrary number of objects. 
We evaluate our method on the task of unsupervised scene decomposition.
Experimental results demonstrate that {\bf SPAIR3D} has strong scalability and is capable of detecting and segmenting an unknown number of objects from a point cloud in an unsupervised manner.

\end{abstract}

\section{Introduction}

3D scenes can exhibit complex and combinatorially large observation spaces even when there are only a few basic elements.
% Robust algorithms are needed to identity objects from 3D observations.
Motivated in part by cognitive psychology studies \cite{ObjectFile} that suggest human brains organize observations at an object level, 
recent advances in physical prediction \cite{OOPhysics1}, and the superior robustness demonstrated by object-oriented reinforcement learning agent\cite{OOMDP,SchemaNetworks}, 
we tackle in this paper the problem of deep object-centric learning on point clouds, which is crucial for high-level relational reasoning and scalable machine intelligence. 

There is a good body of existing literature on unsupervised object-centric generative models for images and videos.
The spatial mixture models are widely adopted to model observations in an object-oriented way~\cite{MONET,SPAIR,AIR,IODINE,SPACE}. 
% SPAIR~\cite{SPAIR} makes use of an object-specification scheme which allows the method to scale well to scenes with a large number of objects. 
These approaches effectively define objectness as a region with strong appearance correlations, and Variational Autoencoders (VAE)~\cite{betaVAE,VAE} play a critical role in exploiting such correlations.
More precisely, the encoder-decoder structure of VAE effectively creates an information bottleneck~\cite{DVIB,UbetaVAE,IB} limiting the amount of information passing through.
To reconstruct the observation under a limited information budget, highly correlated information must be encoded together.
Thus, objectness emerges from the encoding strategy.
The above-mentioned papers mainly exploit appearance correlations on objects that are colored uniformly.
In this paper, we show that this paradigm is also applicable to structural correlations conveyed by point clouds without appearance information, as long as we can overcome some irregularities in point cloud data as described in section~\ref{sec:spair3d}.
% The irregularity of point cloud renders a direct transfer of the above method infeasible (detailed in section \ref{sec:spair3d}).
Specifically, inspired by SPAIR \cite{SPAIR}, we propose in this paper a VAE-based model named
{\bf Sp}atially Invariant {\bf A}ttend, {\bf I}nfer, {\bf R}epeat in {\bf 3D} (SPAIR3D), a model that generates spatial mixture distributions on point clouds to discover 3D objects in static scenes.
Here we summarize the key contributions of this paper:
\begin{itemize}\itemsep1mm\parskip1mm
   \item We propose, to the best of our knowledge, the first unsupervised object-centric learning pipeline for point cloud data, named SPAIR3D.
   \item We also propose a new \emph{Chamfer Mixture Loss} function tailored for learning mixture models over point cloud data with a novel graph neural network that can be used to model and generate a variable number of 3D points.
   \item We provide qualitative and quantitative results to show that SPAIR3D learns meaningful object-centric representation and decomposes point clouds scene with an arbitrary number of objects in an object-oriented manner.
\end{itemize}

\section{Related Work}

\noindent{\bf Generative Unsupervised Object-centric Learning.}  
Unsupervised object-centric learning based on generative models has attracted increasing attention in recent times. 
Such approaches focus on joint object representation-learning and scene decomposition based on single or multiple views~\cite{MONET,ROOTS,SPAIR,GENESIS,IODINE,MulMON,SPACE,ObSuRF}. 
% Spatial Gaussian mixture models are commonly adopted.
A spatial Gaussian mixture model is typically defined on 2D images consisting of $K$ mixture components that correspond to $K$ objects.
Each component spans the entire image and places an isotropic Gaussian on the RGB value of all pixels with predicted mean and a constant covariance.
Each component also assigns each pixel a non-negative mixing weight that sum to one across all components.
The definition can be easily extended to voxel and neural radiance field.

Under the spatial mixture model formulation, different inference methods are proposed.
IODINE~\cite{IODINE} employs iterative amortized inference to refine latent variable posteriors for all components in parallel. 
GENESIS~\cite{GENESIS} and MONET~\cite{MONET} sequentially infer the latent representation, one component at a time.
Slot attention \cite{SlotAttention} and Neural Expectation Maximization (NEM)~\cite{NEM} can be regarded as differentiable clustering algorithms.
% While the inference processes differ, it is shared by the above mentioned works that each mixing component spans across the entire scene.
% The maximum number of possible objects in each scene is set as a hyperparameter.

Instead of treating each component of the mixture model as a full-scale observation, {\bf A}ttend, {\bf I}nfer, {\bf R}epeat (AIR)~\cite{AIR} confines the extent of each component to a local region. 
In AIR, one network is trained to propose a set of candidate object regions in the form of 2D bounding boxes.
Each region is then cropped out and processed by a VAE.
The final reconstruction is obtained by placing the reconstructed patches back in the inferred locations.
% Note that it is equivalent to focusing zero mixing weights outside of the local regions.
Pixels that are not covered by any patches are deemed background.
While AIR fails in scenes of dense objects, SPAIR~\cite{SPAIR} addresses the challenge with a grid spatial attention mechanism with which bounding boxes are proposed locally from each grid cell.
This extension is also proven effective in object-tracking tasks~\cite{SPAIROT}.
By confining the extent of each component, constraints on maximum object sizes are imposed.
SPACE~\cite{SPACE} employs MONET to model background components that are normally much larger than foreground objects.\\
% Spatial attention models are also employed to reconstruct 3D scenes in the form of meshes or voxel in an object-centric fashion from a sequence of RGB frames \cite{ObjectCentricVideoGeneration}. 

\noindent {\bf Graph Neural Network for Point Cloud Generation.}
%Substantial progress has been made in point cloud generation in recent years.
Generative models such as VAEs~\cite{gadelha2018multiresolution} and generative adversarial networks \cite{achlioptas2018learning} have been successfully used for a point-cloud generation but with a pre-defined number of points per object.
It is shown that a point cloud generation process can be modeled as a latent variable conditioned Markov chain \cite{PointMDP}. 
PointFlow \cite{PointFlow} follows a VAE and normalizing flow-based approach that models object shapes as continuous distributions. 
While the proposed approach allows the generation of a variable number of points, it could not be naturally integrated into our framework because of the need for an ODE solver. %due to its requirement of solving an ODE.

\section{SPAIR3D}
\label{sec:spair3d}

% While the literature on deep unsupervised object-centric learning is rich, none of the methods listed above can be applied directly to 3D point cloud data.
While the application of generative-model-based object-centric learning on image~\cite{MONET,IODINE}, voxel~\cite{ObjectCentricVideoGeneration}, and mesh~\cite{ObjectCentricVideoGeneration}
%and neural radiance field (NeRF) data\MM{Not clear what NeRF data is?}~\cite{}
shows encouraging results, its application on point cloud has not been explored up till now.
Unlike point cloud data, the reconstructions of images, and voxels are all coordinate-dependent. % \cite{ObSuRF}\cite{BCVAE}
For each mixture component, given a coordinate, a mixing weight (defining mask) and a feature vector (RGB value) are generated at that coordinate to form a well-defined mixture model.
For image data, the coordinate dependency can be implicitly embedded in the network structure since the input and output are of fixed sizes \cite{CNN}.
The coordinate thus provides the correspondence between input and reconstruction, inducing a natural likelihood function.

However, a point cloud takes the form of an unordered set with some irregular structures.
Each point cloud may have a varying number of points. 
More importantly, the point coordinates carrying structural information becomes the reconstruction target,
% Due to the irregularity of point-cloud data,
there is usually no natural correspondence between the input and the reconstruction.
While~\emph{Chamfer Distance} commonly serves as a loss function for point cloud reconstruction, it does not support mixture model formulation directly.
Such data irregularity makes defining a mixture model over point cloud a non-trivial task. 

To overcome the issues outlined above, we extend the SPAIR framework and introduce {\bf SPAIR3D}, a deep generative model for 3D object-centric learning and 3D scene decomposition via object-centric point-cloud generation.
There are two main reasons for choosing the SPAIR framework over others.
Firstly, as a consequence of the lack of correspondence between input and reconstruction, the likelihood computation commonly involves a bi-directional matching between the generated point cloud and the ground truth, leading to quadratic time complexity. 
The local object proposal and reconstruction mechanism allow us to confine the matching computation in each local region, which significantly reduces the algorithm's time complexity. % (see Supp. Sec. 1 for further analysis).

Secondly, 3D point cloud reconstruction commonly requires the target object to be centered~\cite{PointMDP,PointFlow}.
While it is straightforward to center objects in single-object reconstruction tasks, it is difficult to center all objects in the same coordinate system in the 3D scene reconstruction setup. In contrast, thanks to the local object proposal mechanism proposed in this paper, each object can be naturally centered in a local coordinate system.

In the following, we first introduce our generative model formulation over key latent variables (\S\ref{sec:generative}). 
Then we detail the inference model implementation (\S\ref{sec:inference}). 
We further discuss the particular challenges arising in handling a varying number of points with a novel~\emph{Chamfer Mixture Loss}~(\S\ref{sec:Chamfer}) and~\emph{Point Graph Decoder}~(\S\ref{sec:PGF}). 

\begin{figure}[H]
  \vspace{-5mm}
  \centering
  \begin{subfigure}[b]{0.5\linewidth}
      \centering
      \includegraphics[width=\textwidth]{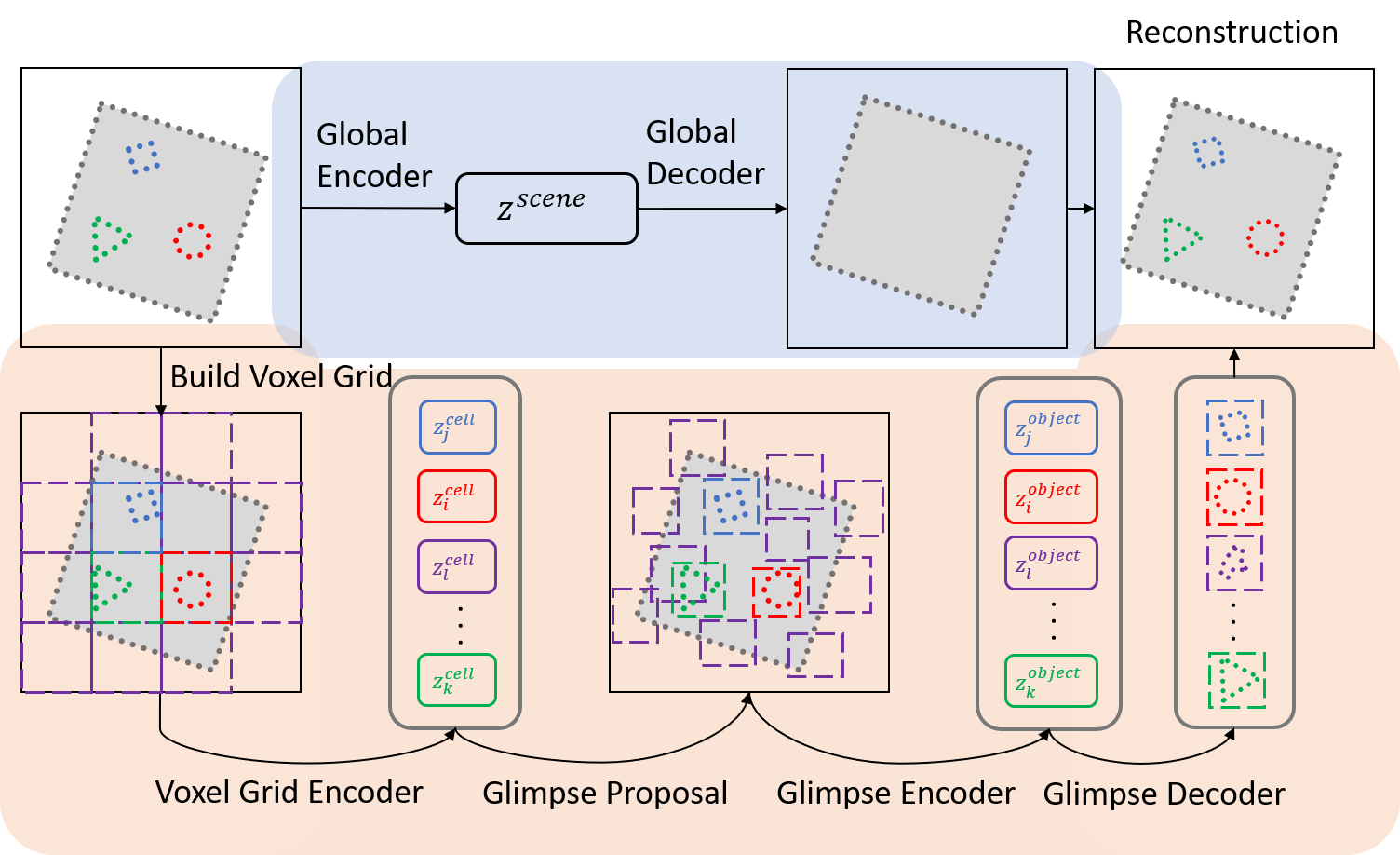}
      \caption{Structure of SPAIR3D}
      \label{fig:spair3d}
  \end{subfigure}
  \hfill
  \begin{subfigure}[b]{0.35\linewidth}
      \centering
      \includegraphics[width=\textwidth]{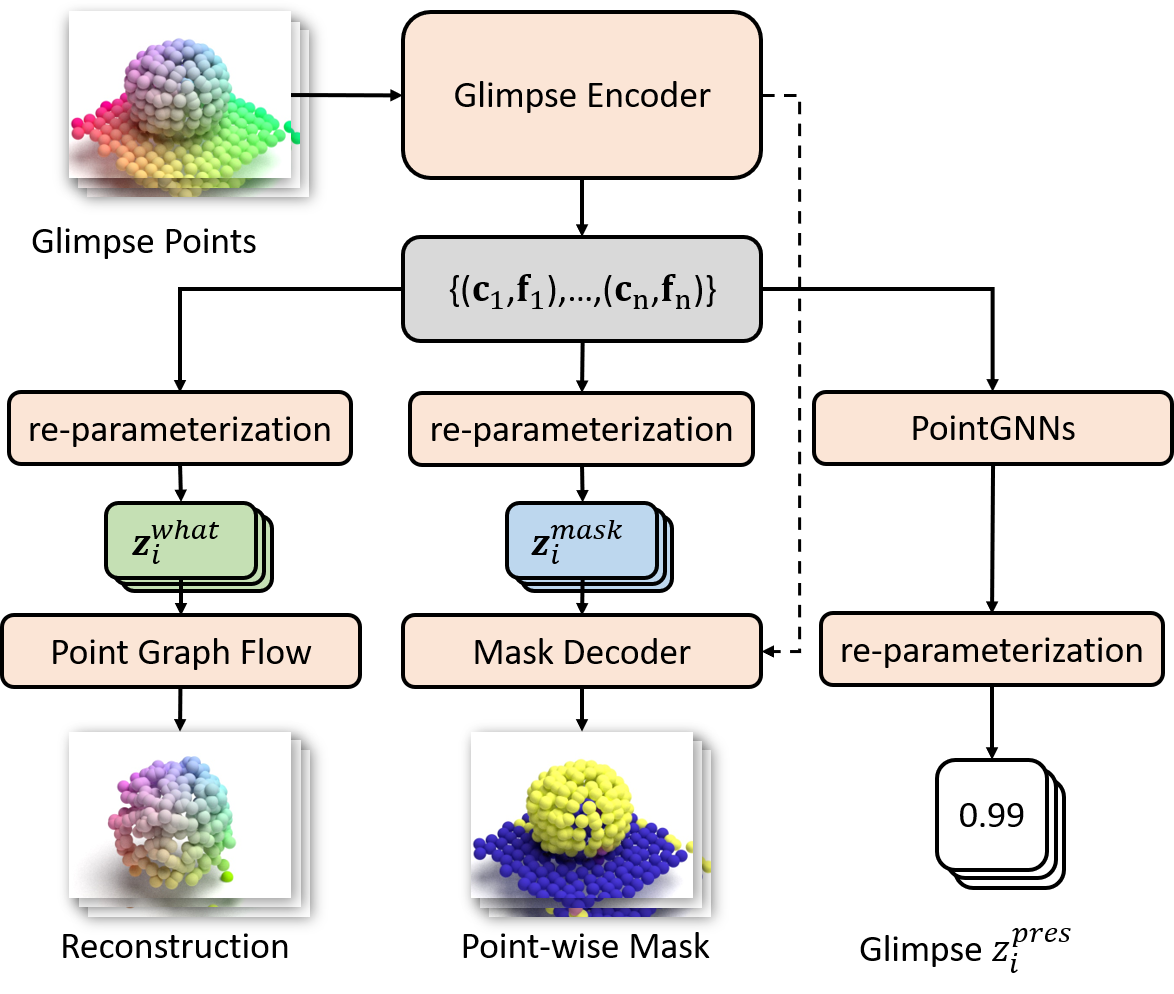}
      \caption{Structure of Glimpse VAE}
      \label{fig:glimpsevae}
  \end{subfigure}
  \caption{(a) Structure of SPAIR3D. For better illustration, we adopt 2D abstraction and use colors to highlight important correspondence.
%    The global VAE branch is highlighted in blue and the foreground object branch is highlighted in orange.
  (b) Structure of Glimpse VAE. 
  Glimpse encoder encodes foreground glimpses and produce $\mathbf{z}^{what}_i$, $\mathbf{z}^{mask}_i$ and $z_i^{pres}$ for each glimpse.
  Point Graph Decoder takes $\mathbf{z}^{what}_i$ and reconstructs input points (left branch).
  Mask Decoder takes $\mathbf{z}^{mask}_i$ and generates masks for each point (middle branch).
  The dashed line represents the dependency on the coordinates of the intermediate points in the hierarchy and $\mathcal{G}_i$.
  Multi-layer PointGNN networks enable message passing between $(\mathbf{c}_i, \mathbf{f}_i)$ and produces $z^{pres}_i$ (right branch).}
  \label{fig:spair3d_}
  \vspace{-5mm}
\end{figure}

\subsection{Local Object Proposal}

As shown in Fig.~\ref{fig:spair3d}, SPAIR3D first divides a 3D scene into a spatial attention voxel grid with possible empty voxel cells covering no points. 
We discard empty cells and associate a bounding box with each non-empty voxel cell.
The set of input points captured by a bounding box is termed an~\emph{object glimpse}. 
Besides object glimpses, SPAIR3D also defines a~\emph{scene glimpse} covering all points in an input scene. 
Later, we show that we encode and reconstruct each glimpse and generate a mixing weight on each point to form a probability mixture model. 

\subsection{Generative Model}\label{sec:generative}
Similar to SPAIR, each grid cell generates posterior distributions over a set of latent variables defined as $\mathbf{z}^{cell}_i = \{\mathbf{z}^{where}_i, \mathbf{z}^{apothem}_i\}$,
where $\mathbf{z}^{where}_i\in \mathbb{R}^3$ encodes the relative position of the center of the $i^{th}$ bounding box to the center of the $i^{th}$ cell, $\mathbf{z}_i^{apothem} \in \mathbb{R}^3$ encodes the apothem of the bounding box.
Thus, each $\mathbf{z}^{cell}_i$ induces one object glimpse associated with the $i^{th}$ cell. 
Each object glimpse is then associated with posterior distributions over latent variables specified as $\mathbf{z}^{object}_i = \{\mathbf{z}^{what}_i, \mathbf{z}^{mask}_i, z^{pres}_i\}$,
where $\mathbf{z}^{what}_i\in \mathbb{R}^A$ encodes the structure information of the corresponding object glimpse, $\mathbf{z}^{mask}_i\in \mathbb{R}^B$ encodes the mask for each point in the glimpse, $z^{pres}_i \in \{0,1\}$ is a binary variable indicating whether the proposed object should exist ($z^{pres}_i$ = 1) or not ($z^{pres}_i$ = 0). 
The scene glimpse is associated with only one latent variable $ \mathbf{z}^{scene} = \{\mathbf{z}^{what}_0\}$.
We assume $z^{pres}_i$ follows a Bernoulli distribution. 
The posteriors and priors of other latent variables are all set to isotropic Gaussian distributions. % (see the Supp. Sec. 2 for details).

Given latent representations of objects and the scene, the complete likelihood for a point cloud $\mathcal{X}$ is formulated as $p(\mathcal{X}) = \int_{\mathbf{z}} p(\mathbf{z}) p(\mathcal{X}|\mathbf{z})d\mathbf{z}$,
where $\mathbf{z} = (\bigcup_i \mathbf{z}^{cell}_i) \cup (\bigcup_i \mathbf{z}^{object}_i) \cup \mathbf{z}^{scene}$.
%~$p(\mathcal{X}|\mathbf{z})$ is the~\emph{Chamfer Mixture Loss} defined below.
As maximizing the objective $p(\mathcal{X})$ is intractable, we resort to the variational inference method to maximize its evidence lower bound (ELBO).

\subsection{Chamfer Mixture Loss} \label{sec:Chamfer}

Unlike generative model-based unsupervised 2D segmentation methods that reconstruct the pixel-wise appearance conditioning on its spatial coordinate, the reconstruction of a point cloud lost its point-wise correspondence to the original point cloud.
~\emph{Chamfer distance} is commonly adopted to measure the discrepancy between the generated point cloud ($\hat{\mathcal{X}}$) and the input point cloud ($\mathcal{X}$).
Formally,~\emph{Chamfer distance} is defined by $d_{CD}(\mathcal{X}, \hat{\mathcal{X}}) = \sum_{x \in \mathcal{X}} \min_{\hat{x}\in \hat{\mathcal{X}}} \norm{x - \hat{x}}_2^2 + \sum_{\hat{x} \in \hat{\mathcal{X}}} \min_{x \in \mathcal{X}} \norm{x - \hat{x}}_2^2$.
We refer to the first and the second term as the forward loss and the backward loss, respectively. 

Unfortunately, the~\emph{Chamfer distance} does not fit naturally into the mixture model framework.
To get around that, we propose a \emph{Chamfer Mixture Loss} (CML) tailored for training probability mixture models defined on point clouds.
The \emph{Chamfer Mixture Loss} is composed of a \emph{forward likelihood} and a \emph{backward regularization} corresponding to the forward and backward loss, respectively.

Denote the $i^{th}$ glimpse as $\mathcal{G}_i$, $i \in \{0, \dots ,n\}$ and its reconstruction as $\hat{\mathcal{G}}_i$, $i \in \{0, \dots ,n\}$.
Specifically, we treat the scene glimpse as the $0^{th}$ glimpse that contains all input points, that is, $\mathcal{G}_0 = \mathcal{X}$.
Note that one input point can be a member of multiple glimpses.
Below we use $\mathcal{N}(x | \mu, \sigma)$ to denote the probability density value of point $x$ evaluated at a Gaussian distribution of mean $\mu$ and variance $\sigma$.
For each input point $x$ in the $i^{th}$ glimpse, the glimpse-wise forward likelihood of that point is defined as $\mathcal{L}_i^F(x) = \frac{1}{u_i}\max_{\hat{x} \in \hat{\mathcal{G}}_i}\mathcal{N}(x | \hat{x}, \sigma_c),$
where $u_i = \int_{x\in \mathcal{X}} \max_{\hat{x} \in \hat{\mathcal{G}}_i}\mathcal{N}(x | \hat{x}, \sigma_c) \mathrm{d}x$ is the normalizer and $\sigma_c$ is a hyperparameter.
For each glimpse $\mathcal{G}_i$, $i \in \{0, \dots ,n\}$,  $\alpha_i^x \in [0, 1]$  defines a mixing weight for point $x$ in the glimpse and $\sum_{i=0}^n\alpha_i^x = 1$. 
In particular, $\alpha_i^x$, $i \in \{1, \dots ,n\}$, is determined by $\alpha_i^x = \frac{z^{pres}_i \pi_i^x}{\sum_{j=1}^n z^{pres}_j  \pi_{j}^x}z^{pres}_i \pi_i^x$,
where $\pi_i^x$ is the predicted mask value and $\pi_i^x =0$ if $x \notin \mathcal{G}_i$.
The mixing weight for the scene layout points completes the distribution through $\alpha_0^x = 1 - \sum_{i = 1}^n \alpha_i^x$ for $x \in \mathcal{G}_0$.
Thus, the final mixture model for an input point $x$ is $\mathcal{L}^F(x) = \sum_{i = 0}^n \alpha_i^x \mathcal{L}_i^F(x).$
The total forward likelihood of $\mathcal{X}$ is then defined as $\mathcal{L}^F(\mathcal{X}) = \prod_{x \in \mathcal{X}} \mathcal{L}^F(x)$.

The forward likelihood alone leads to a trivial sub-optimal solution with $\mathcal{\hat{X}}$ distributed densely and uniformly in the space. 
To enforce a high-quality reconstruction, we define a backward regularization term.
For each predicted point $\hat{x}$, the point-wise backward regularization is $\mathcal{L}^B(\hat{x}) = \max_{x \in \mathcal{G}_{i(\hat{x})}}\mathcal{N}(\hat{x} | x, \sigma_c)$, where $i(\hat{x})$ returns the glimpse index of $\hat{x}$.
We denote $x(\hat{x}) = \arg\max_{x \in \mathcal{G}_{i(\hat{x})}}\mathcal{N}(\hat{x} | x, \sigma_c)$ and $\hat{\mathcal{X}} = \bigcup_{i=0}^n \hat{\mathcal{G}}_i$.
The backward regularization is then defined as $\mathcal{L}^B(\hat{\mathcal{X}}) = \prod_{i=0}^n \prod_{\hat{x} \in \hat{\mathcal{G}}_i} \mathcal{L}^B(\hat{x})^{ \alpha_{i}^{x(\hat{x})}}$.
The exponential weighting, i.e. $\alpha_{i}^{x(\hat{x})}\in [0,1]$, is crucial.
As each predicted point $\hat{x} \in \hat{\mathcal{X}}$ belongs to one and only one glimpse, it is difficult to impose a mixture model interpretation on the backward regularization.
The exponential weighting encourages the generated points in object glimpse to be close to input points with high probability belonging to $\mathcal{G}_i$.
Combining the forward likelihood and the backward regularization together, we define~\emph{Chamfer Mixture Loss} as $\mathcal{L_{CD}}(\mathcal{X}, \hat{\mathcal{X}}) = \mathcal{L}^F(\mathcal{X}) \cdot \mathcal{L}^B(\hat{\mathcal{X}})$.
During inference, the segmentation label for each point $x$ is naturally obtained by $\arg \max_i \alpha_i^x$.

The overall loss function is $\mathcal{L} = -\log\mathcal{L_{CD}}(\mathcal{X}, \hat{\mathcal{X}}) + \mathcal{L}_{KL}(\mathbf{z}^{cell}, \mathbf{z}^{object}, \mathbf{z}^{scene})$, where $\mathcal{L}_{KL}$ is the KL divergence between the prior and posterior of the latent variables. % (Supp. Sec. 2 for details). 
In general, one cannot find a closed-form solution for the normalizer in Chamfer Mixture Loss.
However, the experiments below show that we can safely ignore the normalization constants during optimization.
\vspace{-3mm}

\subsection{Model Structure} \label{sec:modelStr}

We next introduce the encoder and decoder network structure for SPAIR3D. 
% To represent input 3D point clouds, our 
The building blocks are based on graph neural networks and point convolution. % (See Sec. 4 in the Supp. for details).\\

\noindent{\bf Encoder network.}\label{sec:inference}
We design an encoder network $q_\phi({\bf z}|x)$ to obtain the latent representations $\{{\bf z}^{cell}_i\}_{i=1}^n$ and $\{{\bf z}^{object}_i\}_{i=1}^n$ from a point cloud.
%  where $\{{\bf z}^{cell}_i\}_{i=1}^n$ encode information from points in grid cells and $\{{\bf z}^{object}_i\}_{i=1}^n$ encode information for points from object glimpses.
To achieve the spatially invariant property, we group one PointConv~\cite{PointConv} layer and one PointGNN~\cite{PointGNN} layer into pairs for message passing and information aggregation among points and between cells. 
% We now provide details on how we use the encoder for voxel grids and glimpses to learn latent representations.

\noindent{\bf (a) Voxel Grid Encoding}.
The voxel-grid encoder takes a point cloud as input and generates for each spatial attention voxel cell $\mathcal{C}_i$ two latent variables $\mathbf{z}^{where}_i \in \mathbb{R}^3$ and $\mathbf{z}^{apothem}_i \in \mathbb{R}^3$ to propose a glimpse $\mathcal{G}_i$ potentially occupied by an object.  
To better capture the point cloud information in $\mathcal{C}_i$, we build a voxel pyramid within each cell $\mathcal{C}_i$ with the bottom level corresponding to the finest voxel grid. 
We aggregate information hierarchically using PointConv-PointGNN pairs from bottom to top through each level of the pyramid.
%In each layer of the pyramid, all points in each voxel cell are aggregated via PointConv into a new point located at the centroid of the points that are aggregated.
For each layer of the pyramid, we aggregate the features of all points and assign it to the point spawned at the center of mass of the voxel cell. 
Then PointGNN is employed to perform message passing on the radius graph built on all spawned points. 
The output of the final aggregation block produces $\mathbf{z}^{where}_i$ and $\mathbf{z}^{apothem}_i$ via the re-parametrization trick \cite{VAE}.

We obtain the offset distance of a glimpse center from its corresponding grid cell center using $\Delta g_i=\tanh(\mathbf{z}^{where}_i)\cdot L$, where $L$ is the maximum offset distance. 
The apothems of the glimpse in the $x, y, z$ direction is given by $\Delta {\bf g}^{apo}_i =T(\mathbf{z}^{apothem}_i)(\mathbf{r}^{max} - \mathbf{r}^{min}) + \mathbf{r}^{min}$, where % $\Delta {\bf g}^{apo}_i\in \mathbb{R}^3$,
$T(\cdot)$ is the sigmoid function and $[\mathbf{r}^{min}, \mathbf{r}^{max}]$ defines the range of apothem.

\noindent\textbf{(b) Glimpse Encoding.}
The predicted glimpse center offset and the apothems uniquely determine one glimpse for each spatial attention voxel cell. % $\mathcal{C}_i$. 
We adopt the same encoder structure to encode each
glimpse $\mathcal{G}_i$ % independently 
into one point 
$\mathbf{a}_i = (\mathbf{c}_i, \mathbf{f}_i)$, where
$\mathbf{c}_i$ is the glimpse center coordinate and
$\mathbf{f}_i$ is the glimpse feature vector.
% $\mathbf{a}_i$ is 
We then generate $\mathbf{z}^{what}_i$ and $\mathbf{z}^{mask}_i$ from $\mathbf{a}_i$ via the
re-parameterization trick.

% ANCHOR 
% ZPres Generator:
The variable $z^{pres}_i$ governs the glimpse rejection process and is crucial to the final decomposition quality.
Unlike previous work \cite{SPAIR}\cite{SPACE}, SPAIR3D generates
$z^{pres}$ from glimpse features instead of cell features based on
our observation that message passing across glimpses provides more benefits in the glimpse-rejection process. 
To this end, a radius graph is first built on the point set 
$\{(\mathbf{c}_i, \mathbf{f}_i) \}_{i=1}^n$ to connect nearby glimpse centers, which is followed by multiple PointGNN layers with decreasing output channels to perform local message passing. 
The $z^{pres}_i$ of each glimpse is then obtained via the re-parameterization trick. 
% We argue that 
Information exchange between nearby glimpses can help avoid over-segmentation that would otherwise occur because of the high dimensionality of point cloud data.

\noindent\textbf{(c) Global Encoding.} The global encoding module adopts the same encoder as the object glimpse encoder to encode scene glimpse $\mathcal{G}_0$. 
The learned latent representation is $\mathbf{z}^{what}_0$ with $z^{pres}_0 = 1$.\\
% ANCHOR Point Graph Decoder

\noindent{\bf Decoder network.}
% We now introduce the decoders used for point-cloud reconstruction and mask value assignment.
We now introduce the decoders used for point-cloud and mask generation.

\noindent\textbf{(a) Point Graph Decoder (PGD).} \label{sec:PGF}
Given the $\mathbf{z}^{object}_i$ of each glimpse, the decoder is used for point-cloud reconstruction as well as segmentation-mask generation. 
% Most existing decoder or point-cloud generation frameworks can only generate a pre-defined fixed number points, which can lead to under- or over-segmentation.
In reconstruction, the number of generated points has a direct effect on the magnitudes of the forward and backward terms in the Chamfer Mixture Loss.
An unbalanced number of reconstruction points can lead to under- or over-segmentation.
To balance the forward likelihood and the backward regularization, the number of predictions for each glimpse should be approximately the same as the number of input points.
We propose a graph network based point decoder to allow setting the size of $\mathcal{\hat{X}}$ in run time.

PGD treats the point cloud reconstruction as a point diffusion process \cite{PointMDP}.
The input to the PGD is a set of 3D points with coordinates sampled from a zero-centered Gaussian distribution, with the population determined by the number of points in the current glimpse. 
Features of the input points are set uniformly to the latent variable $\mathbf{z}^{what}_i$. PGD is composed of several PointGNN layers, each of which is preceded by a radius graph operation.
~The output of each PointGNN layer is of dimension $f + 3$, with the first $f$ dimensions interpreted as the updated features and the last 3 dimensions interpreted as the updated 3D coordinates for estimated points.
~Since we only focus on point coordinates prediction, we set $f = 0$ for the last PointGNN layer. 

\noindent{\bf (b) Mask Decoder.}
The Mask Decoder decodes $(\mathbf{c}_i, \mathbf{z}^{mask}_i)$ to the mask value, $\pi_i^x \in [0, 1]$, of each point within a glimpse $\mathcal{G}_i$.
The decoding process follows the exact inverse pyramid structure of the Glimpse Encoder. 
To be more precise, the mask decoder can access the spatial coordinates of the intermediate aggregation points of the Glimpse Encoder as well as the point coordinates of $\mathcal{G}_i$.
During decoding, PointConv is used as deconvolution operation.

\noindent{\bf Glimpse VAE and Global VAE.}
The complete Glimpse VAE structure is presented in Fig.~\ref{fig:glimpsevae}.
The Glimpse VAE is composed of a Glimpse Encoder, Point Graph Decoder, Mask Decoder and a multi-layer PointGNN network.
The Glimpse Encoder takes all glimpses as input and encodes each glimpse $\mathcal{G}_i$ individually and in parallel into feature points $(\mathbf{c}_i, \mathbf{f}_i)$.
Via the re-parameterization trick, $\mathbf{z}^{what}_i$ and $\mathbf{z}^{mask}_i$ are then obtained from $\mathbf{f}_i$.
From there, we use the Point Graph Decoder to decode
~$\mathbf{z}^{what}_i$ to reconstruct the input points, and we use the Mask Decoder to decode $\mathbf{z}^{mask}_i$ to assign a mask value for each input point within $\mathcal{G}_i$.
Finally, $\mathbf{z}^{pres}_i$ is generated via message passing among neighbour glimpses. 
All glimpses are processed in parallel.
The Global VAE consisting of the Global Encoder and a PGD outputs the reconstructed scene layout.
\vspace{-3mm}

\subsection{Soft Boundary}

The prior of $\mathbf{z}^{apothem}$ is set to encourage apothem to shrink so that the size of the glimpses will not be overly large. 
However, if points are excluded from one glimpse, the gradient from the likelihood of the excluded points will not influence the size and location of the glimpse anymore, and this can lead to over-segmentation.
To solve this problem, we introduce a soft boundary weight $b^x_i \in [0, 1]$ which decreases when a point $x \in \mathcal{G}_i$ moves away from the bounding box of $\mathcal{G}_i$.
Taking $b^x_i$ into the computation of $\alpha$, we obtain an updated mixing weight $\overline{\alpha}_i^x = \frac{z^{pres}_i \pi_i^x b^x_i}{\sum_{j=1}^n z^{pres}_j  \pi_{j}^x b^x_j}z^{pres}_i \pi_i^x b^x_i$.
By employing such a boundary loss, the gradual exclusion of points from glimpses will be reflected in gradients to counter over-segmentation. 
% Details can be found in Supp. Sec. 3.
\vspace{-3mm}

\section{Experiments}

\subsection{Simulated Datasets}

\noindent \textbf{Dataset Generation.}
While many benchmark datasets have been established \cite{CLEVR}\cite{multiobjectdatasets19} for unsupervised object-centric learning, they do not come in the form of a point cloud.
Thus, we introduce two new point-cloud datasets {\it Unity Object Room} and {\it Unity Object Table} built on the~\emph{Unity} platform~\cite{Unity}. 
The Unity Object Room (UOR) dataset is built to approximate the Object Room \cite{multiobjectdatasets19} dataset but with increased scope and complexity.
In each scene, objects sampled from a list of $8$ regular geometries are randomly placed on a square floor.
The Unity Object Table (UOT) dataset approximates the Robotic Object Grasping scenario where multiple objects are placed on a round table.
Instead of using objects of simple geometries, we populate each scene with objects from a pool of $9$ objects with challenging irregular structures.
For both datasets, the number of objects placed in each scene varies from $2$ to $5$ with equal probabilities. 
During the scene generation, the size and orientation of the objects are varied randomly within a pre-defined range. 

We capture the depth, RGB, normal frames, and pixel-wise semantics as well as instance labels for each scene from 10 different viewpoints. 
This setup aims to approximate the scenario where a robot equipped with depth and RGB sensors navigates around target objects and captures data. 
The point cloud data for each scene is then constructed by merging these $10$ depth maps.
For each dataset, we collect $50K$ training scenes, $10K$ validation scenes and $5K$ testing scenes.
% In-depth dataset specification and analysis can be found in Supp. Sec. 5.

% \subsubsection{Unsupervised Segmentation}

\noindent \textbf{Baseline.}
Due to the sparse literature on unsupervised 3D point cloud object-centric learning, we could not find a generative baseline to compare with.
Thus, we compare SPAIR3D with PointGroup (PG) \cite{PointGroup}, a recent supervised 3D point cloud segmentation model.
PointGroup is trained with ground-truth semantic labels and instance labels and performs semantic prediction and instance predictions on a point cloud.
To ensure a fair comparison, we assign each point the same color (white).
The PointGroup network is fine-tuned on the validation set to achieve the best performance.

\noindent \textbf{Performance Metric.}
% For quantitative measurement, 
For UOR and UOT dataset, we use the Adjust Rand Index (ARI)~\cite{ARI} to measure the segmentation performance against the ground truth instance labels.
We also employ foreground Segmentation Covering (SC) and foreground unweighted mean Segmentation Covering (mSC) \cite{GENESIS} for performance measurements as ARI does not penalize object over-segmentation \cite{GENESIS}.
% Forward (CD F) and backward (CD B) Chamfer distance between the input and our reconstruction are also reported. 

\begin{figure}[ht]
  \vspace{-0.5cm}
  \centering
  \begin{subfigure}[b]{0.225\linewidth}
      \centering
      \includegraphics[width=\textwidth]{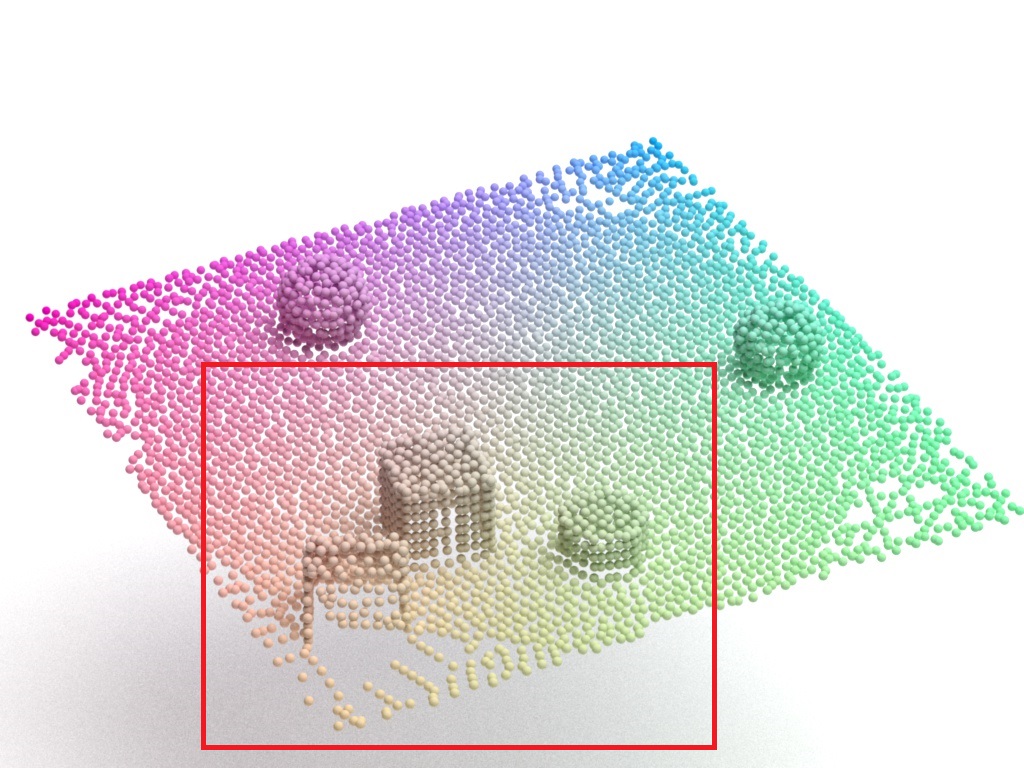}
      \caption{UOR input}
      \label{fig:scene_gt_UOR}
  \end{subfigure}
  \hfill
  \begin{subfigure}[b]{0.225\linewidth}
      \centering
      \includegraphics[width=\textwidth]{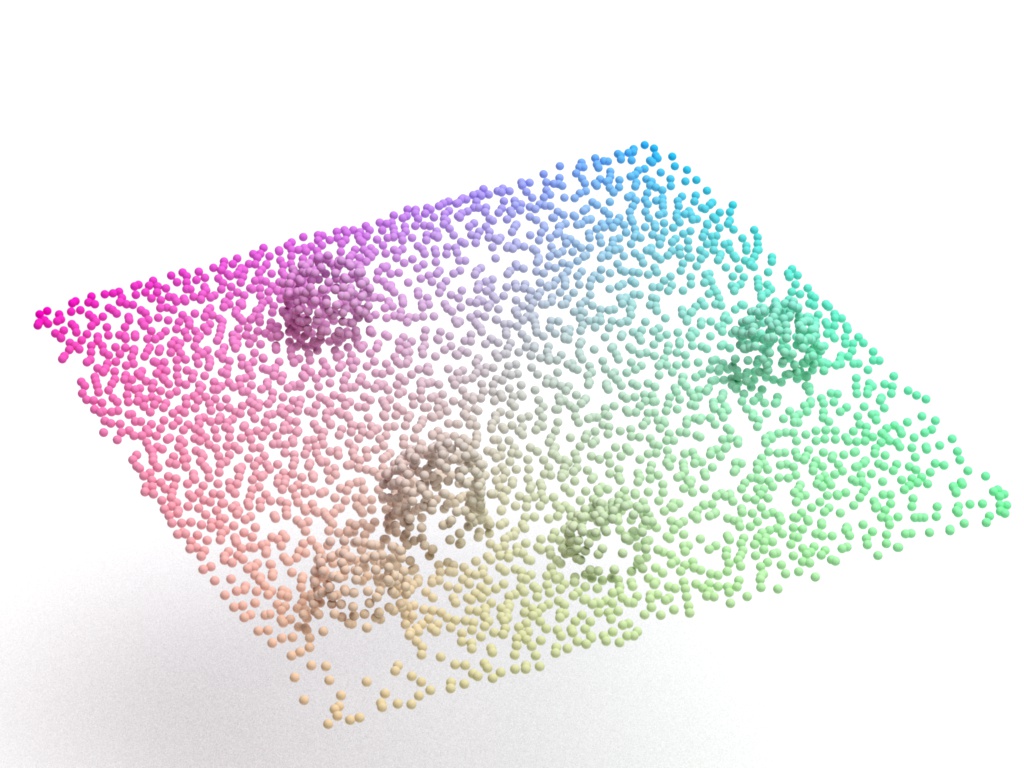}
      \caption{reconstruction}
      \label{fig:scene_Recon_UOR}
  \end{subfigure}
  \hfill
  \begin{subfigure}[b]{0.225\linewidth}
     \centering
     \includegraphics[width=\textwidth]{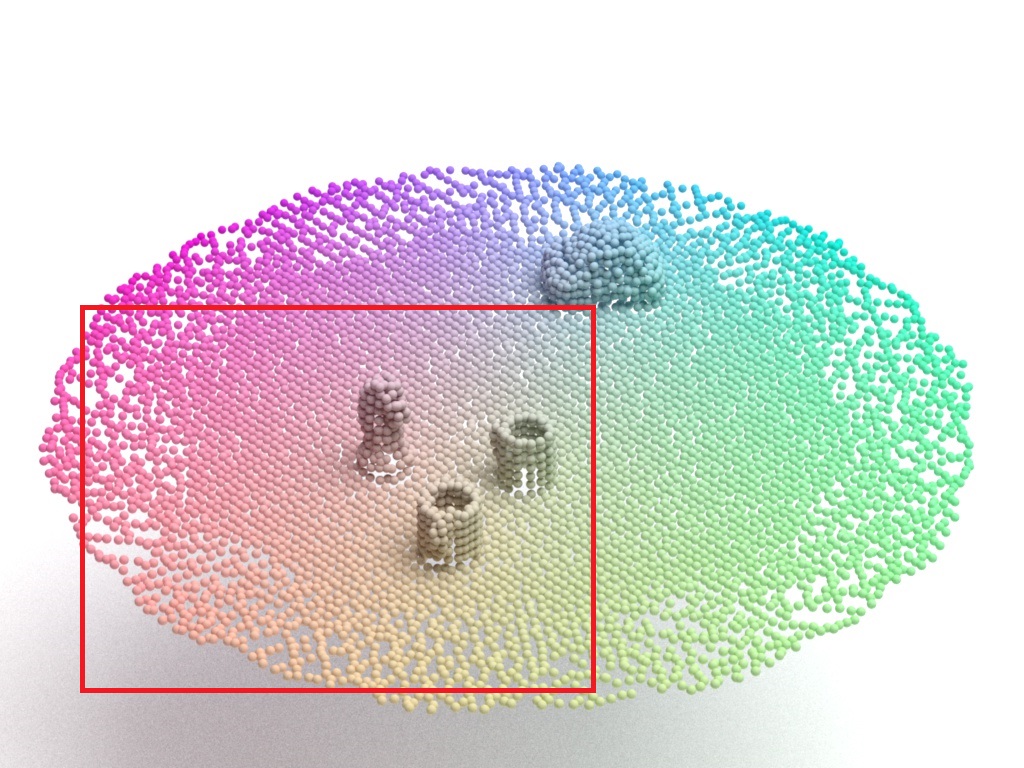}
     \caption{UOT input}
     \label{fig:scene_gt_UOT}
  \end{subfigure}
  \hfill
  \begin{subfigure}[b]{0.225\linewidth}
        \centering
        \includegraphics[width=\textwidth]{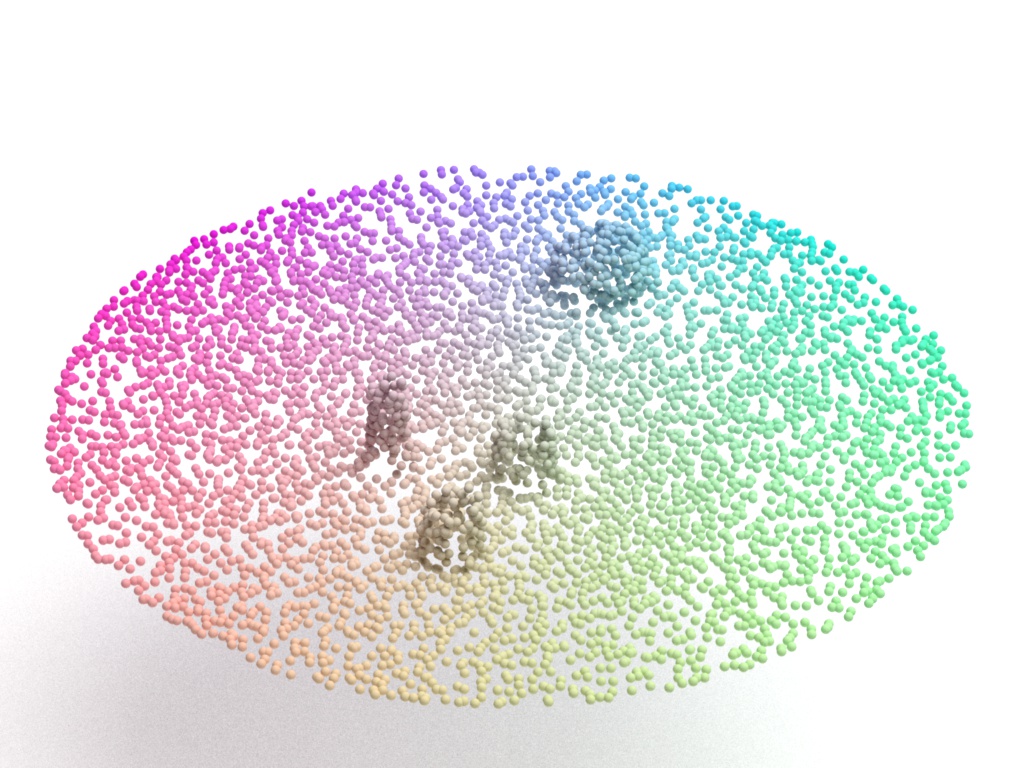}
        \caption{reconstruction}
        \label{fig:scene_Recon_UOT}
  \end{subfigure}
  \hfill
  \begin{subfigure}[b]{0.225\linewidth}
      \centering
      \includegraphics[width=\textwidth]{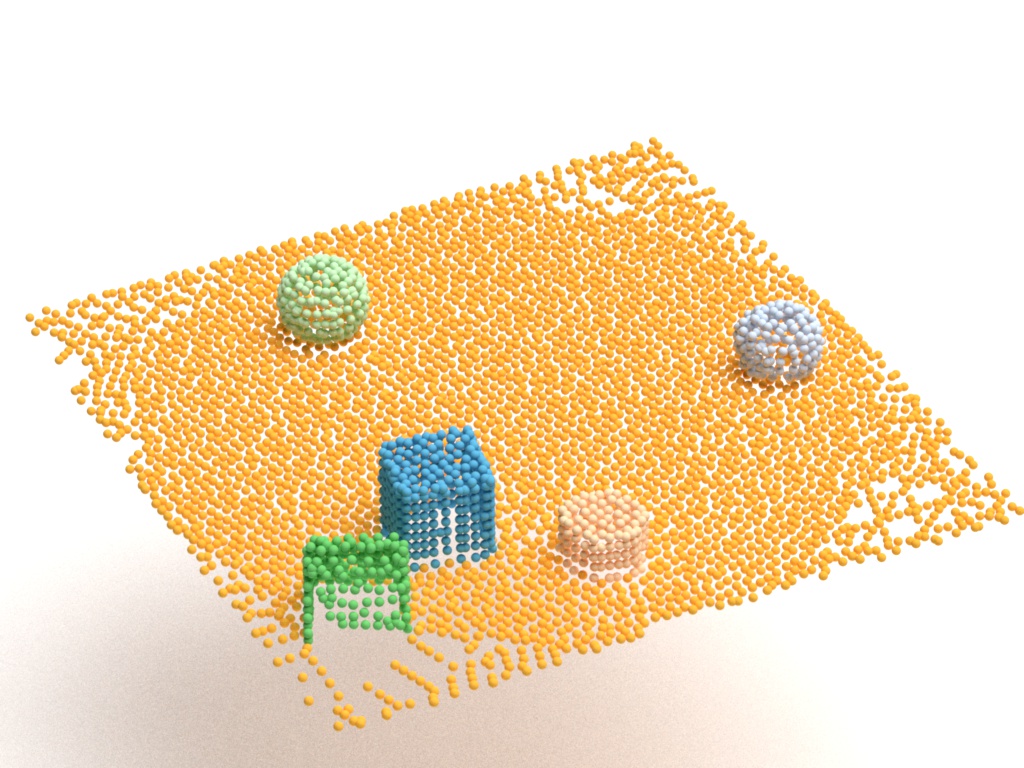}
      \caption{instance label}
      \label{fig:scene_Id_UOR}
  \end{subfigure}
  \hfill
  \begin{subfigure}[b]{0.225\linewidth}
      \centering
      \includegraphics[width=\textwidth]{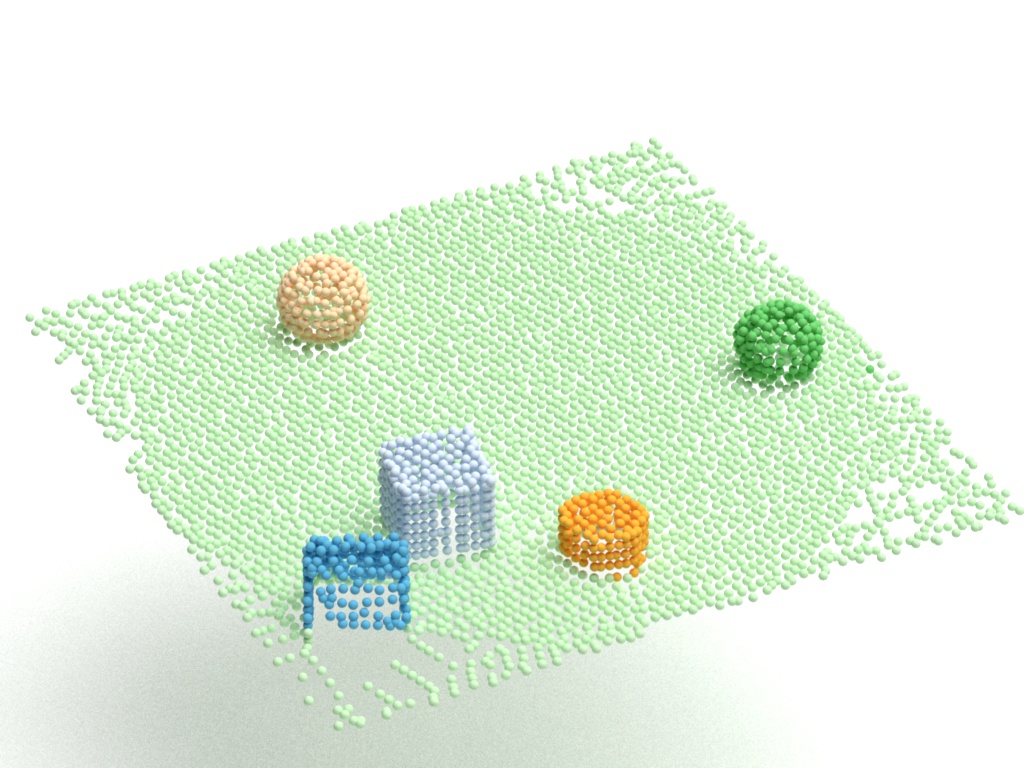}
      \caption{segmentation}
      \label{fig:scene_Seg_UOR}
  \end{subfigure}
  \hfill
  \begin{subfigure}[b]{0.225\linewidth}
     \centering
     \includegraphics[width=\textwidth]{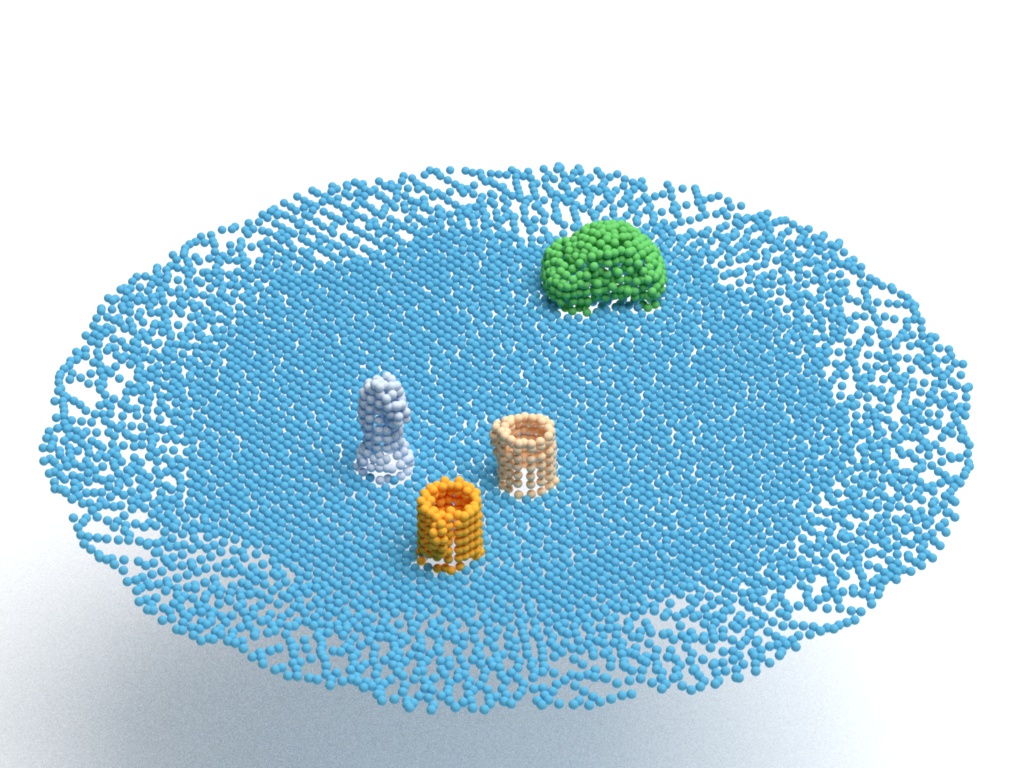}
     \caption{instance label}
     \label{fig:scene_Id_UOT}
  \end{subfigure}
  \hfill
  \begin{subfigure}[b]{0.225\linewidth}
        \centering
        \includegraphics[width=\textwidth]{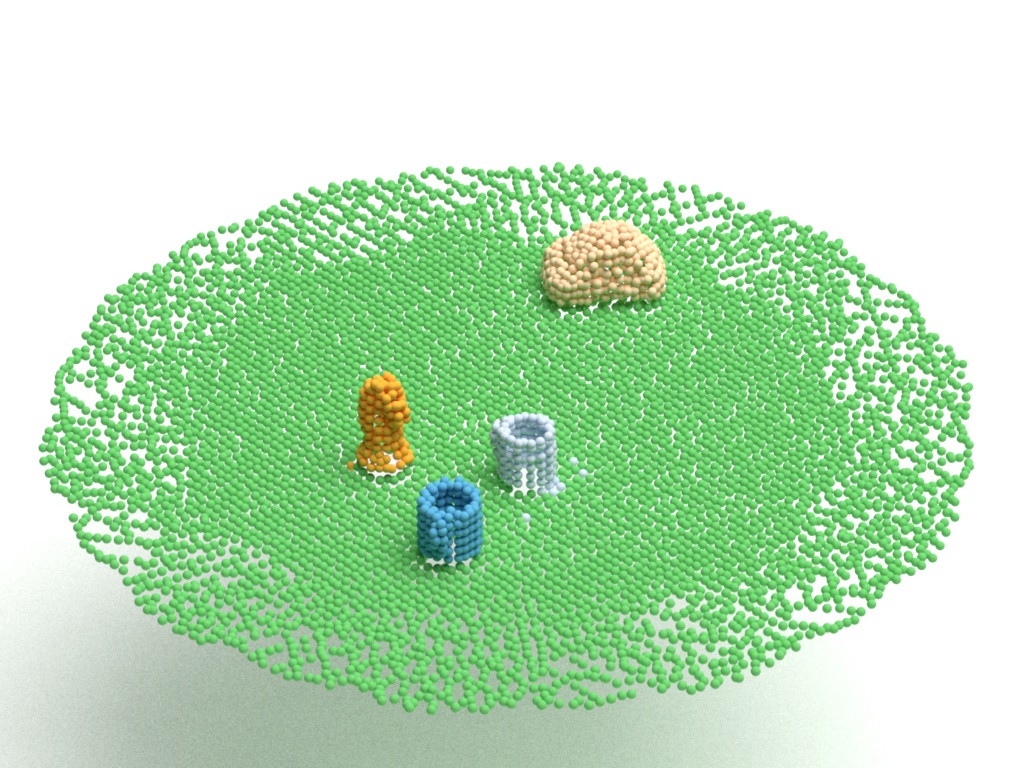}
        \caption{segmentation}
        \label{fig:scene_Seg_UOT}
  \end{subfigure}
  \hfill
  \begin{subfigure}[b]{0.45\linewidth}
     \centering
     \includegraphics[width=\textwidth]{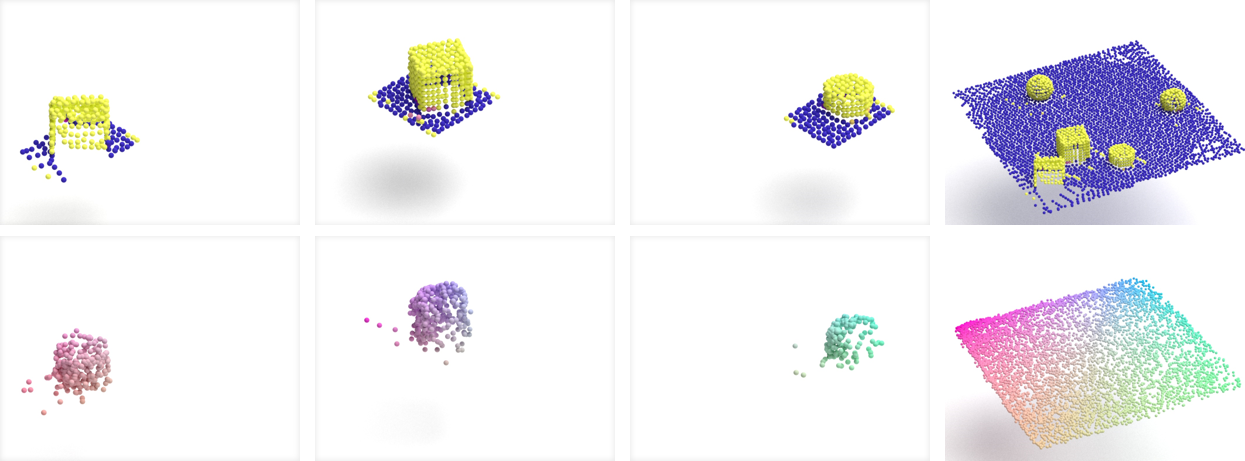}
     % \caption{Visualization of each glimpse (the first and the third row) and the corresponding reconstruction (the second and fourth row). The bottom right image demonstrates the reconstruction result of the global VAE.}
     \caption{UOR close-up glimpses visualization, foreground alpha and scene layout reconstruction.}
     \label{fig:comp_UOR}
 \end{subfigure}
  \hfill
  \begin{subfigure}[b]{0.45\linewidth}
     \centering
     \includegraphics[width=\textwidth]{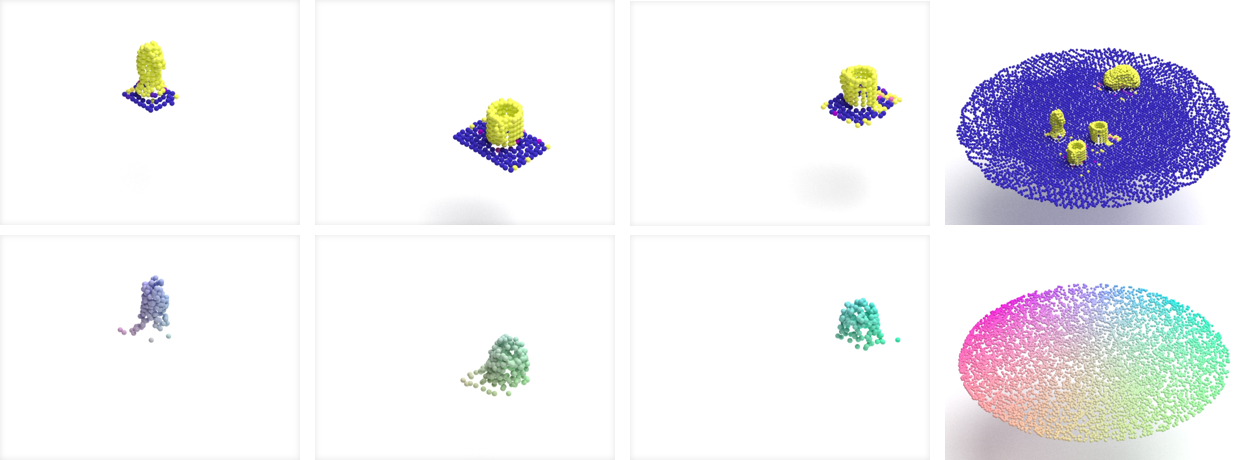}
     % \caption{Visualization of each glimpse (the first and the third row) and the corresponding reconstruction (the second and fourth row). The bottom right image demonstrates the reconstruction result of the global VAE.}
     \caption{UOT close-up glimpses visualization, foreground alpha and scene layout reconstruction.}
     \label{fig:comp_UOT}
  \end{subfigure}
     \caption{Visualization of segmentation results on UOR and UOT dataset.}
     \label{fig:SPAIR3D}
     \vspace{-6mm}
\end{figure}

\noindent \textbf{Evaluation.}
Table \ref{table:results} shows that SPAIR3D achieves comparable performance to the supervised baseline on both UOT and UOR datasets.
As demonstrated in Fig.~\ref{fig:SPAIR3D}, each foreground object is proposed by one and only one glimpse.
The scene layout is separated from objects and accurately modelled by the global VAE.
It is worth noting that the segmentation errors mainly happen at the bottom of objects.
Without appearance information, points at the bottom of objects are also correlated to the ground.
% Since the prior of $\mathbf{z}^{cell}$ encourages glimpses to shrink, those points are likely segmented as part of the scene layout instead of objects.
In Fig.~\ref{fig:test_distribution}, we sort the test data based on their performance in ascending order and plot the performance distributions.
As expected, the supervised baseline (Orange) performs better but SPAIR3D manages to achieve high-quality segmentation (SC score $>0.8$) on around $80\%$ of the scenes without supervision. 
% See Supp. Sec. 6 for more segmentation results.

\begin{figure}[H]
  \vspace{-5mm}
  \begin{subfigure}[b]{0.2\linewidth}
    \centering
    \includegraphics[width=\textwidth]{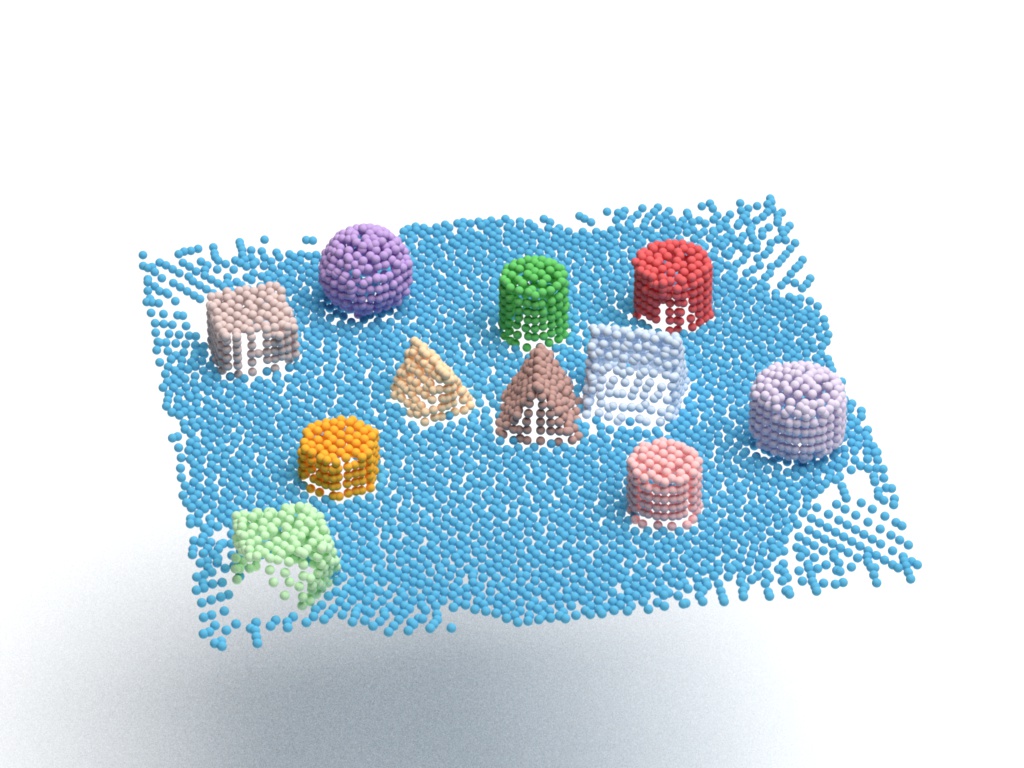}
    % \caption{instance label}
    %  \label{fig:scene_gt_UOT}
  \end{subfigure}
  \hfill
  \begin{subfigure}[b]{0.2\linewidth}
    \centering
    \includegraphics[width=\textwidth]{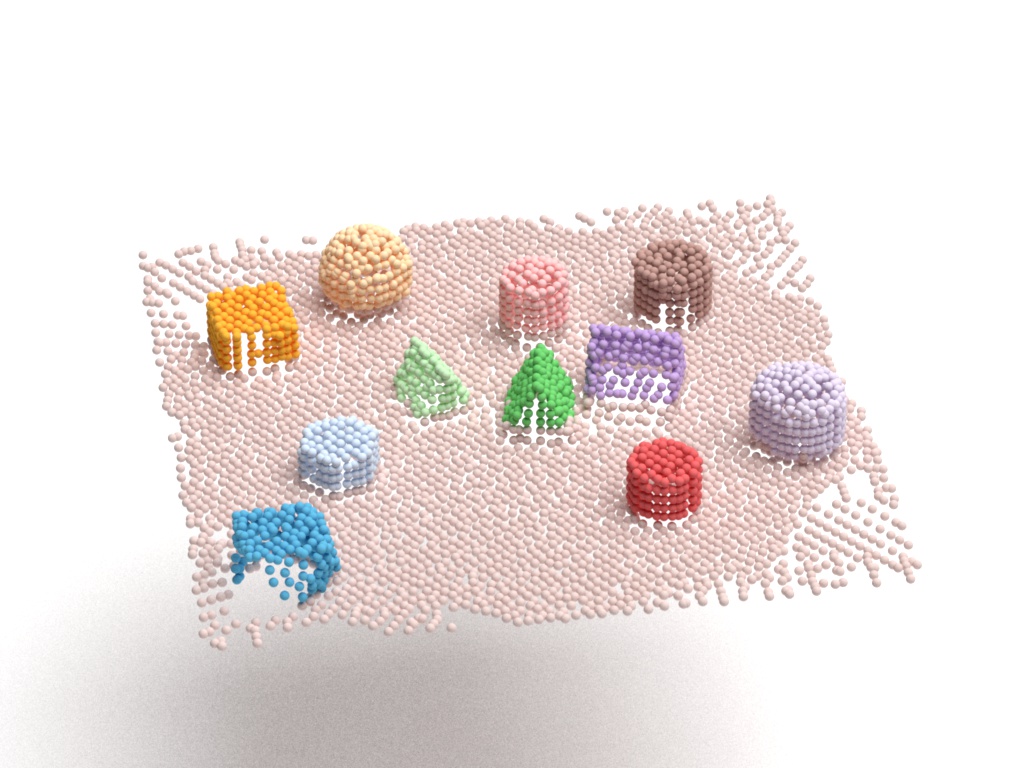}
    % \caption{segmentation}
    %  \label{fig:scene_Recon_UOT}
  \end{subfigure}
  \hfill
  \begin{subfigure}[b]{0.2\linewidth}
    \centering
    \includegraphics[width=\textwidth]{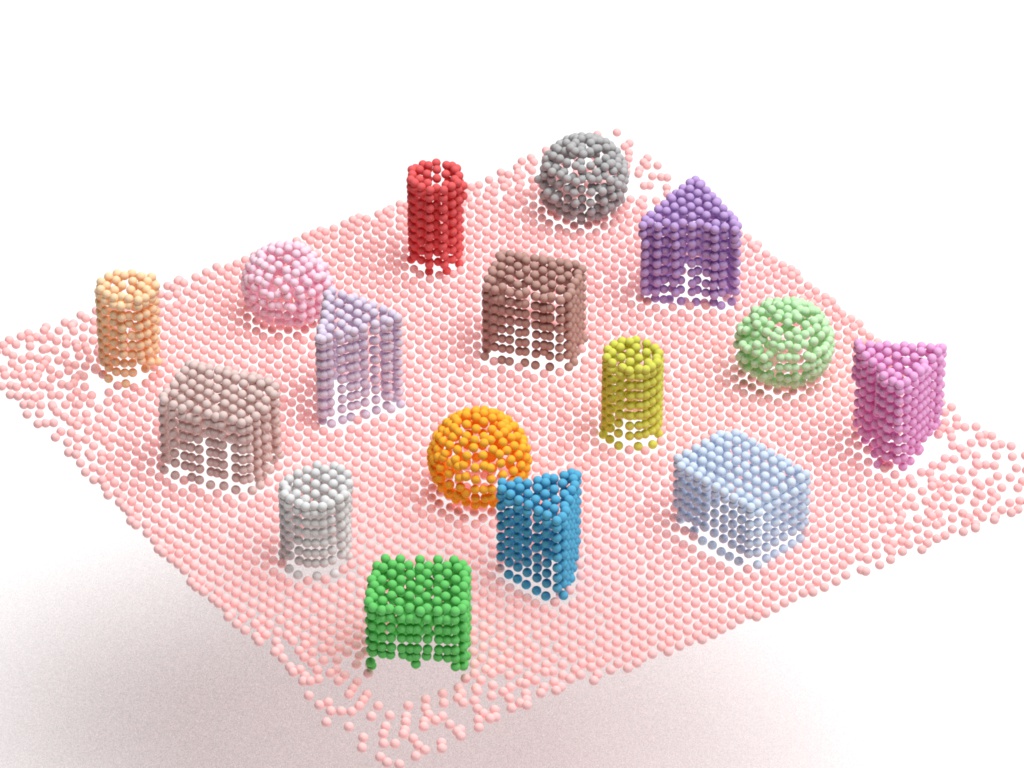}
    % \caption{instance label}
  %  \label{fig:scene_gt_UOT}
  \end{subfigure}
  \hfill
  \begin{subfigure}[b]{0.2\linewidth}
    \centering
    \includegraphics[width=\textwidth]{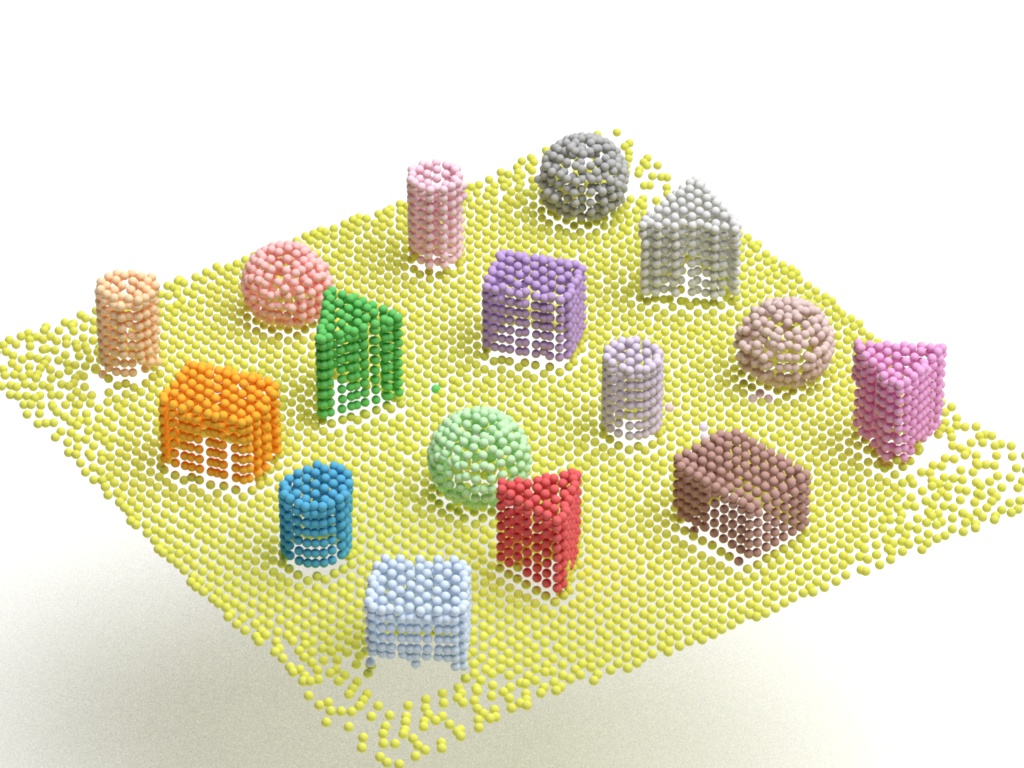}
    % \caption{segmentation}
    %  \label{fig:scene_Recon_UOT}
  \end{subfigure}
  \hfill
  \begin{subfigure}[b]{0.2\linewidth}
    \centering
    \includegraphics[width=\textwidth]{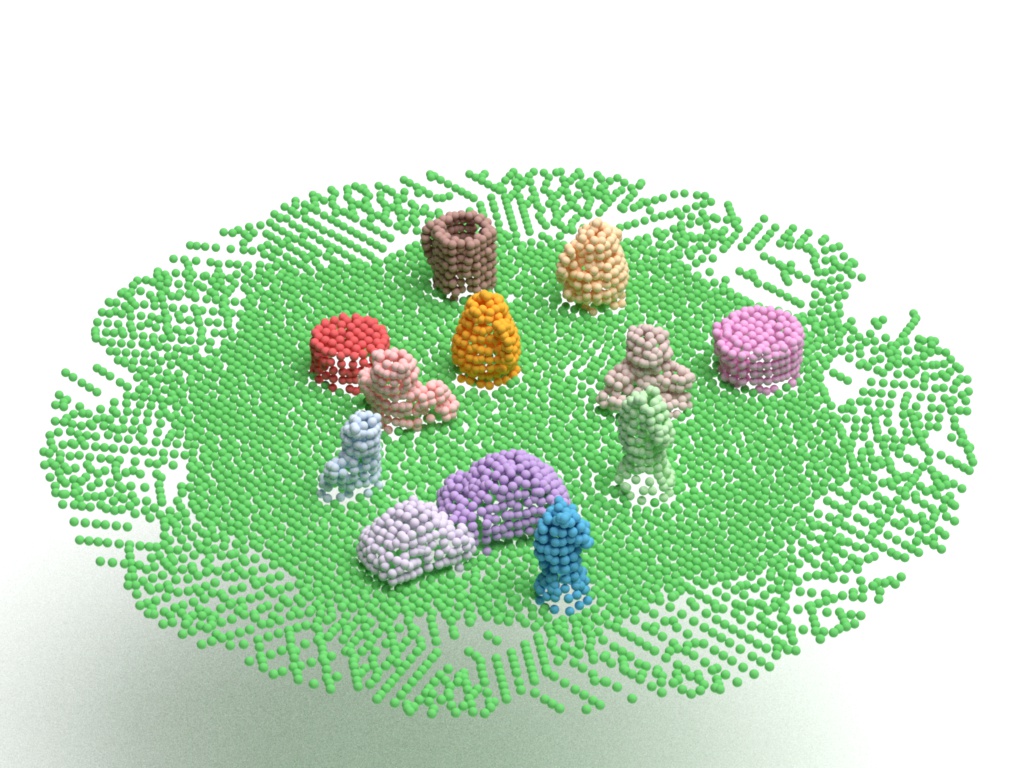}
    \caption{instance label}
  %  \label{fig:scene_Id_UOT}
  \end{subfigure}
  \hfill
  \begin{subfigure}[b]{0.2\linewidth}
    \centering
    \includegraphics[width=\textwidth]{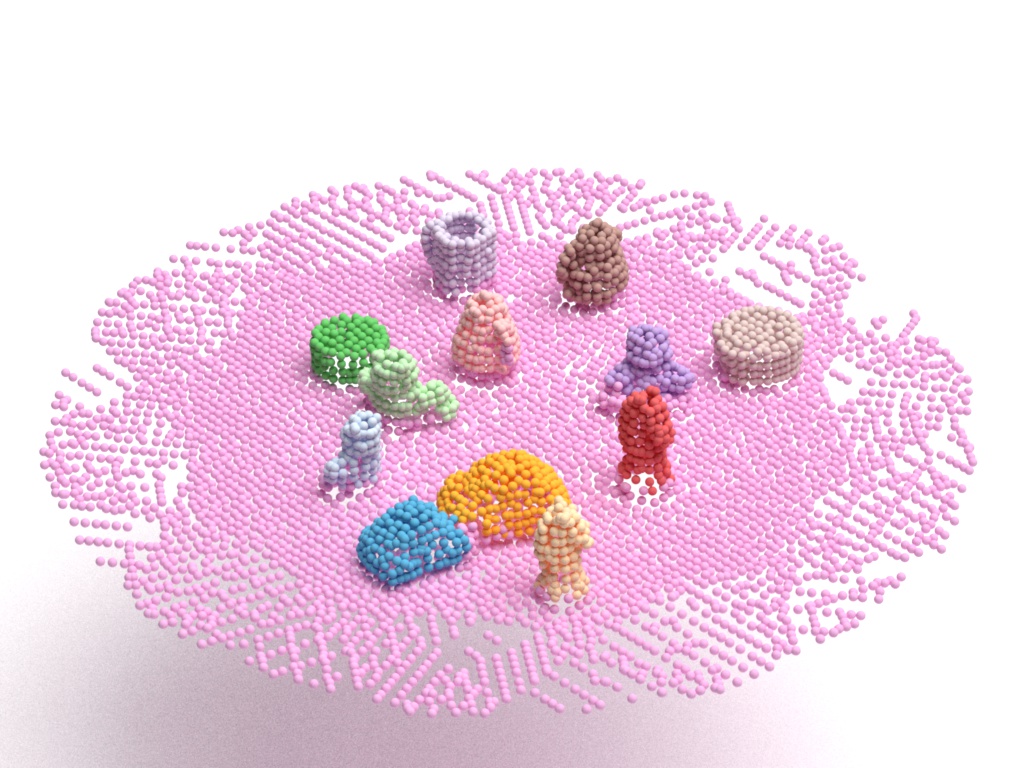}
    \caption{segmentation}
    %  \label{fig:scene_Seg_UOT}
  \end{subfigure}
  \hfill
  \begin{subfigure}[b]{0.2\linewidth}
    \centering
    \includegraphics[width=\textwidth]{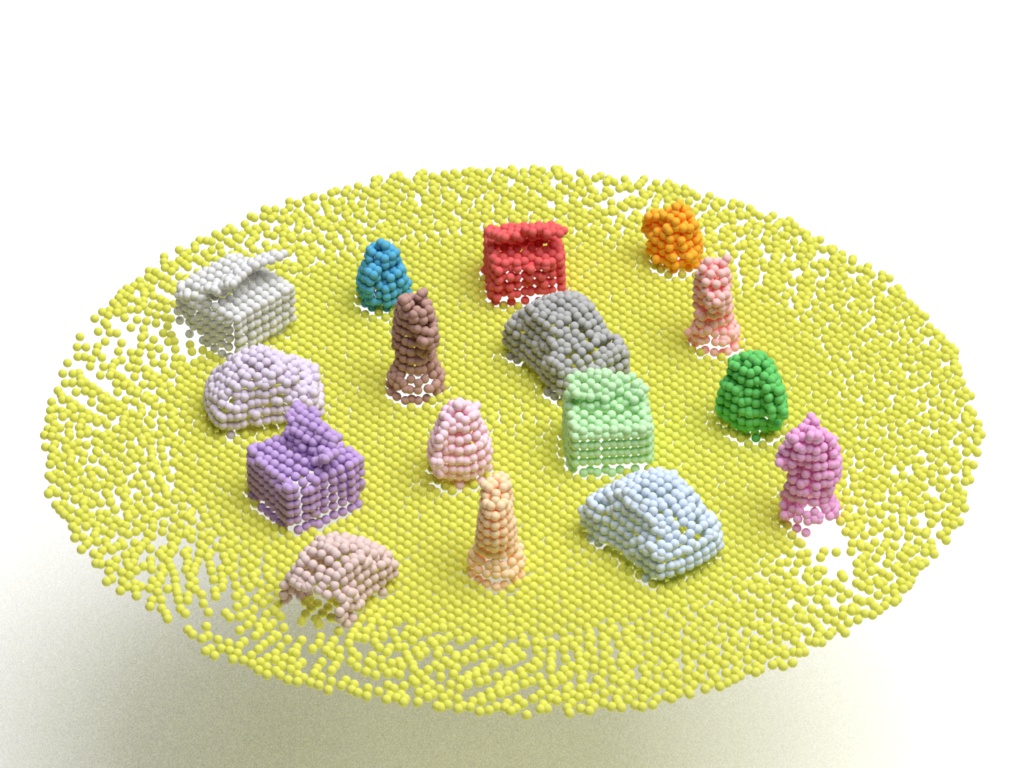}
    \caption{instance label}
    %  \label{fig:scene_Id_UOT}
  \end{subfigure}
  \hfill
  \begin{subfigure}[b]{0.2\linewidth}
    \centering
    \includegraphics[width=\textwidth]{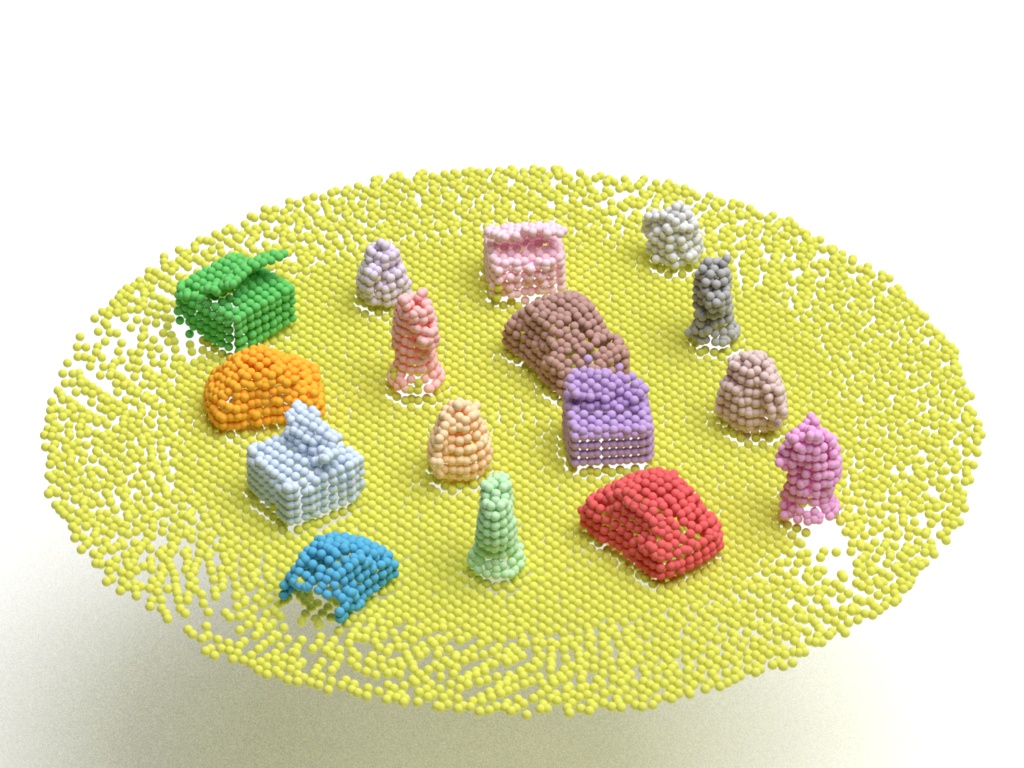}
    \caption{segmentation}
    %  \label{fig:scene_Seg_UOT}
  \end{subfigure}
  \centering
  \caption{Segmentation on 6 - 12 object scenes (a,b) and object matrix (c,d)}
  \label{fig:scale}
\end{figure}

It is worth noting that UOR and UOT datasets include around $25\%$ objects (for scenes of 2 to 5 objects) and $60\%$ objects (for scenes of 6 to 12 objects) that are close to its neighbors with touching or almost touching surfaces. 
The reported quantitative (Table~\ref{table:results}) and qualitative results~(Fig.~\ref{fig:scale}(a)--(d)) show that our method achieves stable performance for those challenging scenes.

\begin{table}[H]
  % \vspace{-2cm}
  \begin{center}
  \begin{tabular}{c|c|c|c|c|c}
  UOR \cellcolor{blue!25}                 &  \multirow{2}{*}{PG}        & \multirow{2}{*}{SPAIR3D (ours)}                &  voxel size 0.75$l$          & \multirow{2}{*}{$6-12$ objects} & \multirow{2}{*}{object matrix} \\ \cline{1-1}
  UOT \cellcolor{red!25}                  &                             &                                      &  voxel size 1.25$l$          &                                 & \\ \hline
  \multirow{2}{*}{ARI$\uparrow$}          & $0.976$ \cellcolor{blue!25} & $0.915 \pm 0.03$ \cellcolor{blue!25} & $0.932$ \cellcolor{blue!25}  & $0.912$ \cellcolor{blue!25} & $0.872$ \cellcolor{blue!25}  \\
                                          & $0.923$ \cellcolor{red!25}  & $0.901 \pm 0.02$ \cellcolor{red!25}  & $0.922$ \cellcolor{blue!25}  & $0.892$ \cellcolor{red!25}  & $0.879$ \cellcolor{red!25}   \\ \hline
  \multirow{2}{*}{SC$\uparrow$}           & $0.907$ \cellcolor{blue!25} & $0.832 \pm 0.04$ \cellcolor{blue!25} & $0.853$ \cellcolor{blue!25}  & $0.846$ \cellcolor{blue!25} & $0.856$ \cellcolor{blue!25}  \\
                                          & $0.917$ \cellcolor{red!25}  & $0.835 \pm 0.03$ \cellcolor{red!25}  & $0.857$ \cellcolor{blue!25}  & $0.843$ \cellcolor{red!25}  & $0.877$ \cellcolor{red!25}   \\ \hline
  \multirow{2}{*}{mSC$\uparrow$}          & $0.900$ \cellcolor{blue!25} & $0.836 \pm 0.04$  \cellcolor{blue!25} & $0.850$ \cellcolor{blue!25}  & $0.842$ \cellcolor{blue!25} & $0.861$ \cellcolor{blue!25} \\
                                          & $0.907$ \cellcolor{red!25}  & $0.831 \pm 0.03$  \cellcolor{red!25}  & $0.861$ \cellcolor{blue!25}  & $0.834$ \cellcolor{red!25}  & $0.886$ \cellcolor{red!25}  \\ \hline
  \end{tabular}     
  \vspace{5mm}   
  \caption{3D point cloud segmentation results on UOR (blue) and UOT (red).}
  \label{table:results}
  \vspace{-2cm}
\end{center}
\end{table}

\subsubsection{Object Centric Representation}
\begin{figure}
  \centering
  \includegraphics[width=0.5\linewidth]{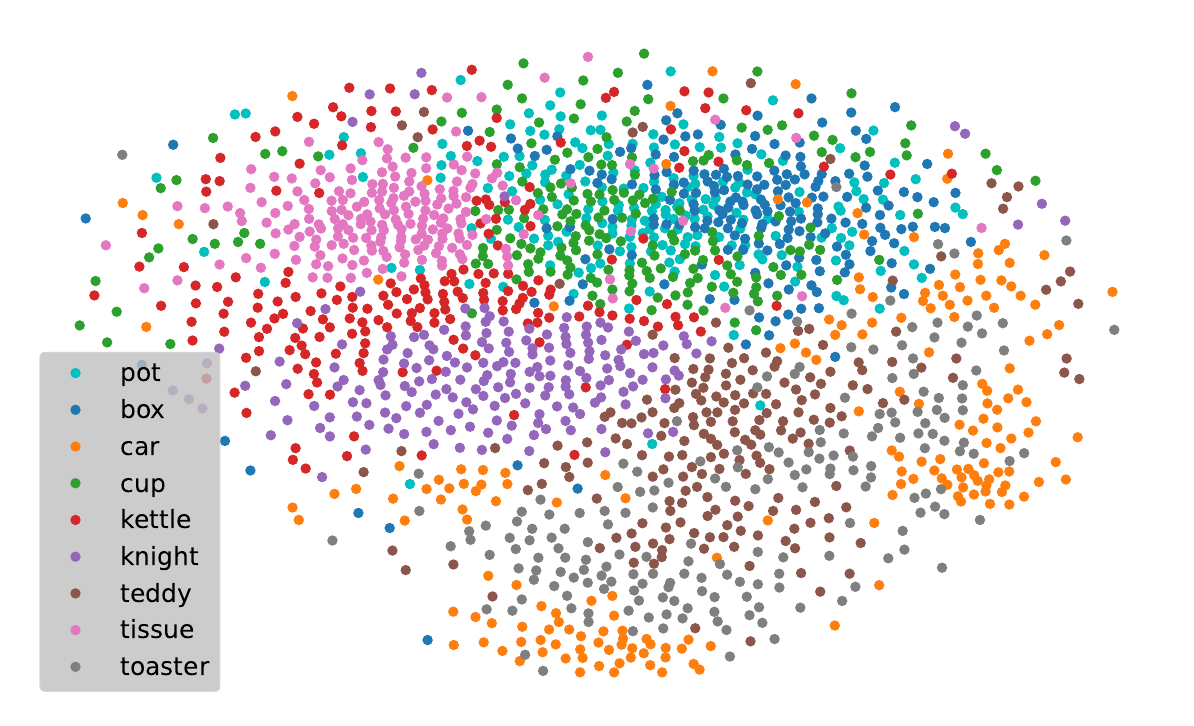}
  \caption{t-SNE visualization of $z^{what}$ on UOT}
  \label{fig:tsne}
  \vspace{-1cm}
\end{figure}

One advantage of our model is simultaneous segmentation and representation learning.
To show that our model learns meaningful representations, we collect the $z^{what}$ of $200$ instances per-type and visualize them with the t-SNE algorithm \cite{tSNE}.
Fig~\ref{fig:tsne} shows the $z^{what}$ of different object types occupy different regions.
Note the embeddings of~\emph{pot} and~\emph{box} instances occupy the same area since they have almost identical spatial structure.
\vspace{-1cm}

\subsubsection{Voxel Size Robustness and Scalability} \label{sec:size}

In the literature, the cell voxel size, an important hyperparameter, is chosen to match the object size in the scene \cite{SPAIR}\cite{SPACE}.
To evaluate the robustness of our method w.r.t voxel size, we train our model on the UOR dataset with voxel size set to $0.75l$ and $1.25l$ with $l$ being the average size of the objects.
Results in Table~\ref{table:results} show that our method
achieves stable performance w.r.t the voxel size.
%We note that the performance is stable w.r.t the voxel size (see Table~\ref{table:results2}).

To demonstrate scalability, we evaluate our pre-trained model on $1000$ scenes containing $6-12$ randomly selected objects and report performance in Table~\ref{table:results}.
Due to the spatial invariance property, SPAIR3D suffers no performance drop on $6-12$ object scenes that were never used for training.

We also evaluated our approach on scenes termed as \textit{Object Matrix}, which consists of $16$ objects placed in a matrix form.
We fixed the position of all $16$ objects but set their size and rotation randomly.
For each dataset, SPAIR3D is evaluated on $100$ \textit{Object Matrix} scenes.
The results are reported in Table~\ref{table:results}.
Note that our model is trained on scenes with $2$ to $5$ objects, which is less than one-third of the number of objects in \textit{Object Matrix} scenes.
Fig~\ref{fig:scale}(c)-(d) is illustrative of the results.

\begin{figure*}
	\centering
	\begin{tabular}{ccc}
		\includegraphics[width=0.3\textwidth]{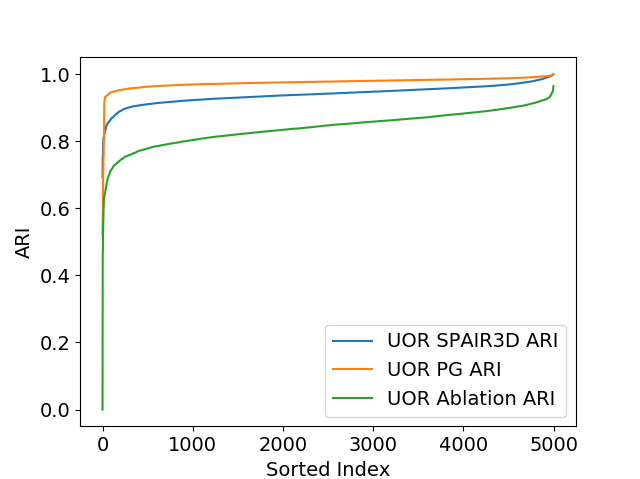}&  \includegraphics[width=0.3\textwidth]{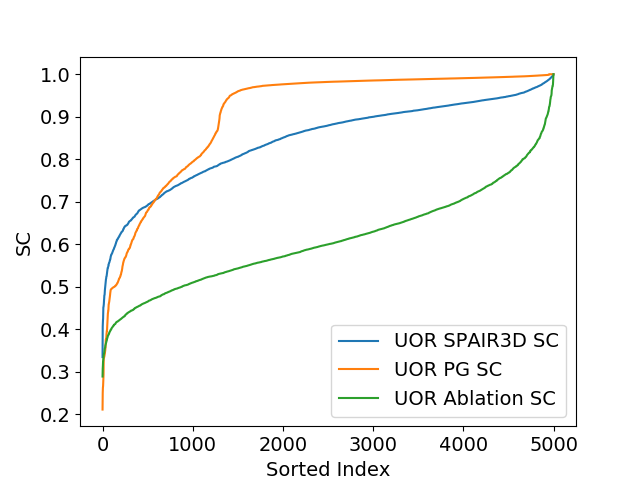}& \includegraphics[width=0.3\textwidth]{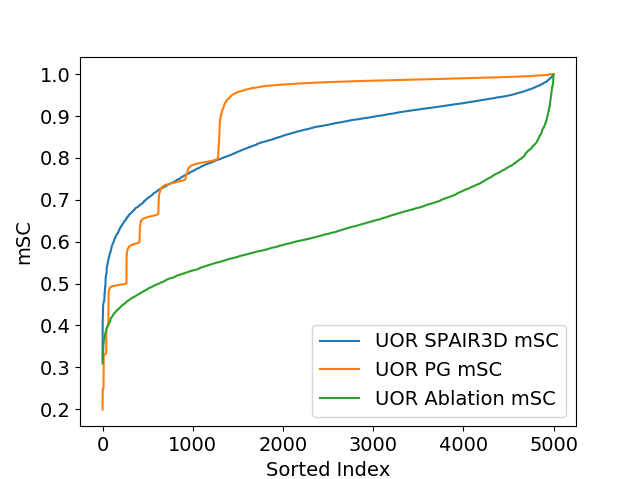}\\
		% (a)&(b)&(c)\\
        \includegraphics[width=0.3\textwidth]{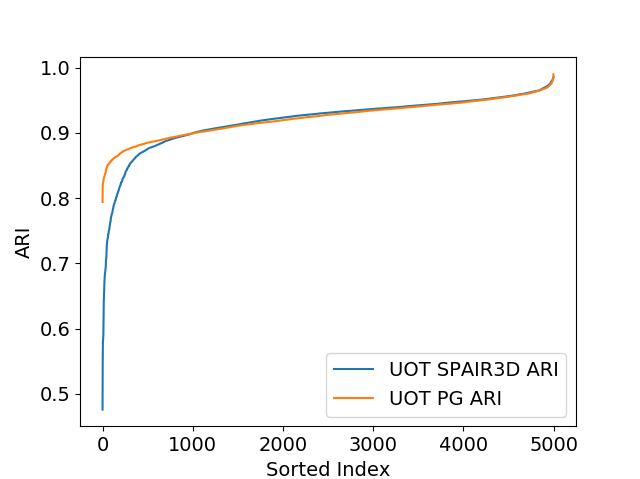}&  \includegraphics[width=0.3\textwidth]{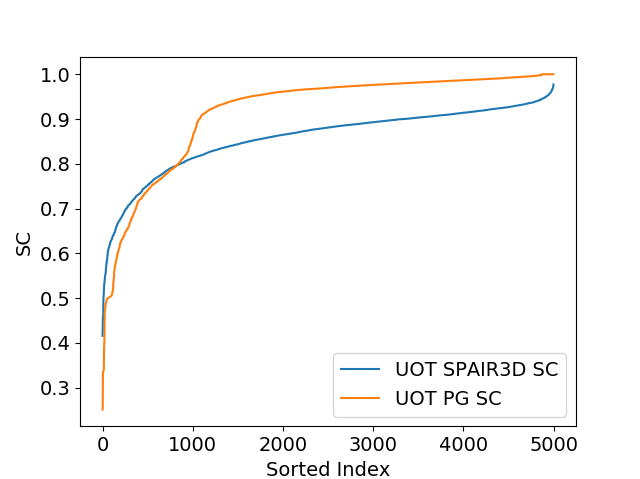}&   \includegraphics[width=0.3\textwidth]{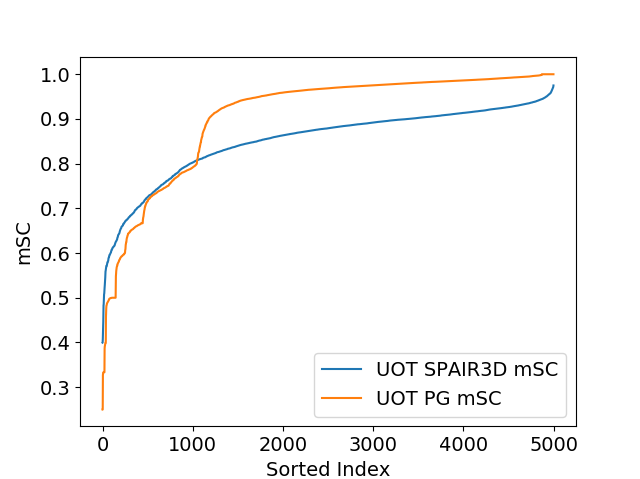}\\
		% (d)&(e)&(f)\\
	\end{tabular}
    % \vspace{0.2cm}
	\caption{Performance distributions on UOR (row one) and UOT (row two).}
	\label{fig:test_distribution}
  \vspace{-5mm}
\end{figure*}

\subsection{Real Dataset}

To demonstrate the performance of our approach on real data,  we apply our model on the S3DIS~\cite{S3DIS} dataset, which contains point clouds of 6 large-scale indoor areas with 271 rooms (scenes).

\noindent \textbf{Data Preprocessing.}
While the dataset contains objects from 13 semantic categories, we focus on objects with regular structures, including chairs, tables, and sofas.
As our approach focuses on object-centric learning, we thus manually inspect the dataset and remove rooms that are too empty (such as hallway), containing clutter (such as storage room), and connected objects (such as lecture theater with connected chairs). 
Finally, we kept 174 scenes in total. We then downsample the dense point cloud of each scene for computational efficiency.
% TODO: S3DIS dataset
% Note that generative unsupervised object-centric learning models require a large amount of training data. 
% For example, IODINE \cite{IODINE} is trained on CLEVR \cite{CLEVR} dataset that consists of 100K images. 
% MONET~\cite{MONET} is trained on Object Room datasets with 1M scenes.
% In contrast, S3DIS dataset only consists of 271 rooms/scenes, which is relatively small.

\noindent \textbf{Baseline.}
Besides Point Group \cite{PointGroup} as a supervised baseline, we additionally use Mean-shift as a rule-based as well as an unsupervised baseline.
Floors are manually removed before applying Mean-shift and the bandwidth parameter is determined by grid-search. 

\noindent \textbf{Performance Metric.}
Due to the scene diversity, instead of reporting the average per-scene ARI, SC, or mSC, we report the per-class mIoU on test sets.
For our model and Point Group, we perform 6-fold cross-validation on the 6 areas and report the average.

\begin{figure*}[ht]
  \centering
  \begin{subfigure}[b]{0.18\linewidth}
      \centering
      \includegraphics[width=\textwidth]{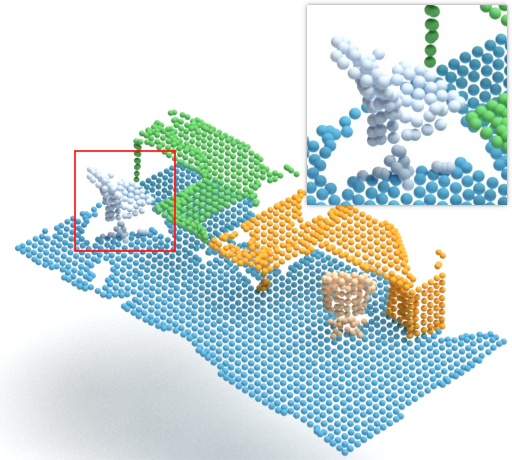}
      % \caption{Instance label}
      % \label{fig:boundary_structure}
  \end{subfigure}
  \hfill
  \begin{subfigure}[b]{0.18\linewidth}
      \centering
      \includegraphics[width=\textwidth]{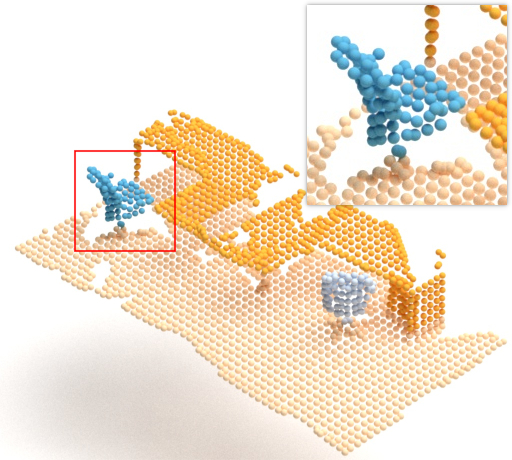}
      % \caption{Instance label}
      % \label{fig:boundary_structure}
  \end{subfigure}
  \hfill
  \begin{subfigure}[b]{0.18\linewidth}
      \centering
      \includegraphics[width=\textwidth]{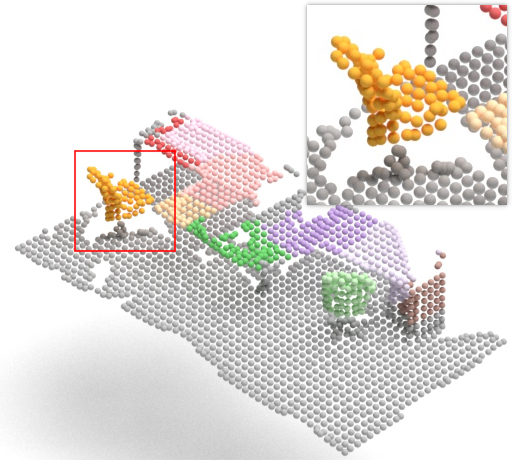}
      % \caption{Our segmentation}
      % \label{fig:boundary_structure}
  \end{subfigure}
  \hfill
  \begin{subfigure}[b]{0.18\linewidth}
      \centering
      \includegraphics[width=\textwidth]{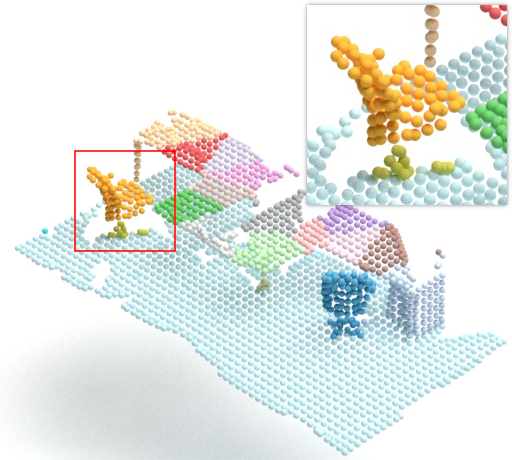}
      % \caption{Instance label}
      % \label{fig:boundary_structure}
  \end{subfigure}
  \hfill
  \begin{subfigure}[b]{0.18\linewidth}
      \centering
      \includegraphics[width=\textwidth]{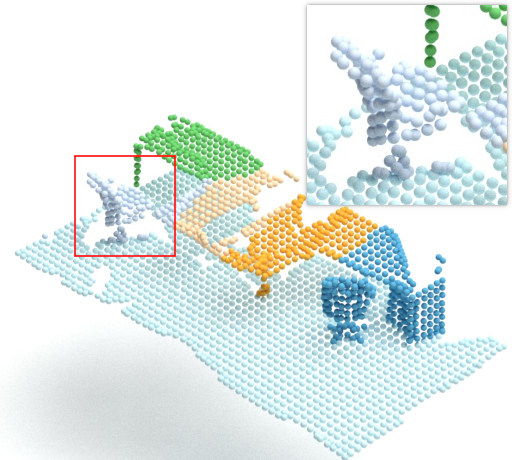}
      % \caption{Our segmentation}
      % \label{fig:boundary_structure}
  \end{subfigure}
  %%%%%%%%%%%%%%%%%%%%%%%%%%%%%%%%%%%%%%%%%%%%%%%%%%%%%%%%%%%%%%%%%%%%%%%%%%%%%%%%%
  % \begin{subfigure}[b]{0.18\linewidth}
  %   \centering
  %   \includegraphics[width=\textwidth]{figures/S3DIS_selected/dash_37_Id__00.jpg}
  %   % \caption{Instance label}
  %   % \label{fig:boundary_structure}
  % \end{subfigure}
  % \hfill
  % \begin{subfigure}[b]{0.18\linewidth}
  %     \centering
  %     \includegraphics[width=\textwidth]{figures/S3DIS_selected/dash_37_Seg__00.jpg}
  %     % \caption{Our segmentation}
  %     % \label{fig:boundary_structure}
  % \end{subfigure}
  % \hfill
  % \begin{subfigure}[b]{0.18\linewidth}
  %     \centering
  %     \includegraphics[width=\textwidth]{figures/S3DIS_MS/37_MS_006_Seg__00.jpg}
  %     % \caption{Instance label}
  %     % \label{fig:boundary_structure}
  % \end{subfigure}
  % \hfill
  % \begin{subfigure}[b]{0.18\linewidth}
  %     \centering
  %     \includegraphics[width=\textwidth]{figures/S3DIS_MS/37_MS_015_Seg__00.jpg}
  %     % \caption{Our segmentation}
  %     % \label{fig:boundary_structure}
  % \end{subfigure}
  % \hfill
  % \begin{subfigure}[b]{0.18\linewidth}
  %     \centering
  %     \includegraphics[width=\textwidth]{figures/S3DIS_PG/train_157_Seg__00.jpg}
  %     % \caption{Instance label}
  %     % \label{fig:boundary_structure}
  % \end{subfigure}
  %%%%%%%%%%%%%%%%%%%%%%%%%%%%%%%%%%%%%%%%%%%%%%%%%%%%%%%%%%%%%%%%%%%%%%%%%%%%%%%%%
  \begin{subfigure}[b]{0.18\linewidth}
    \centering
    \includegraphics[width=\textwidth]{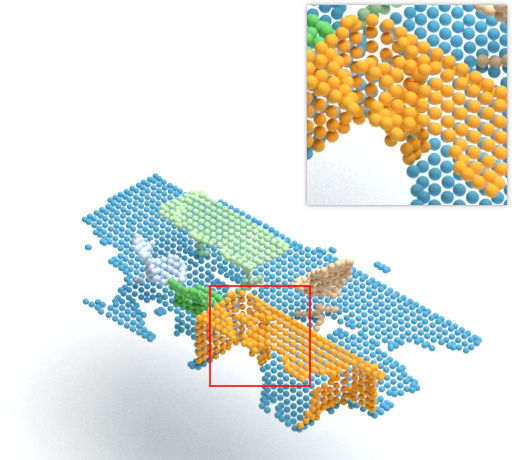}
    % \caption{Instance label}
    % \label{fig:boundary_structure}
  \end{subfigure}
  \hfill
  \begin{subfigure}[b]{0.18\linewidth}
    \centering
    \includegraphics[width=\textwidth]{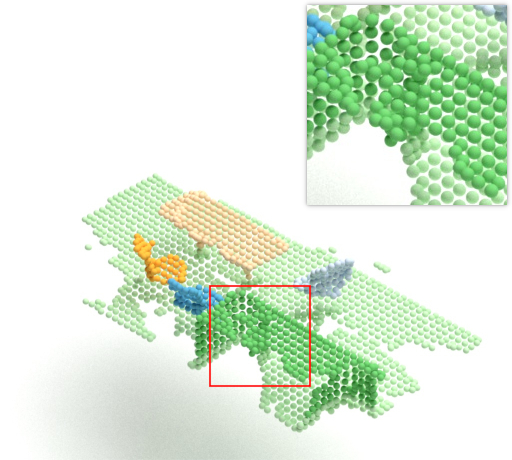}
    % \caption{Instance label}
    % \label{fig:boundary_structure}
  \end{subfigure}
  \hfill
  \begin{subfigure}[b]{0.18\linewidth}
    \centering
    \includegraphics[width=\textwidth]{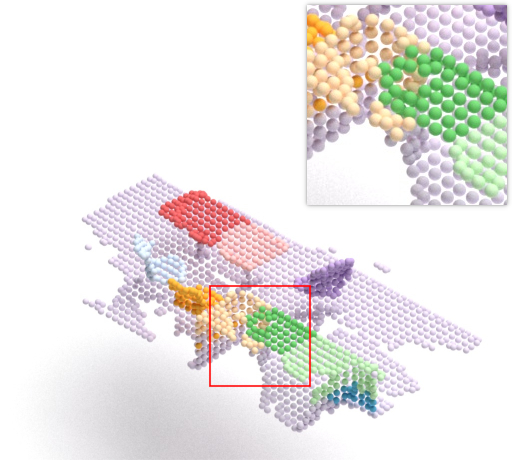}
    % \caption{Our segmentation}
    % \label{fig:boundary_structure}
  \end{subfigure}
  \hfill
  \begin{subfigure}[b]{0.18\linewidth}
    \centering
    \includegraphics[width=\textwidth]{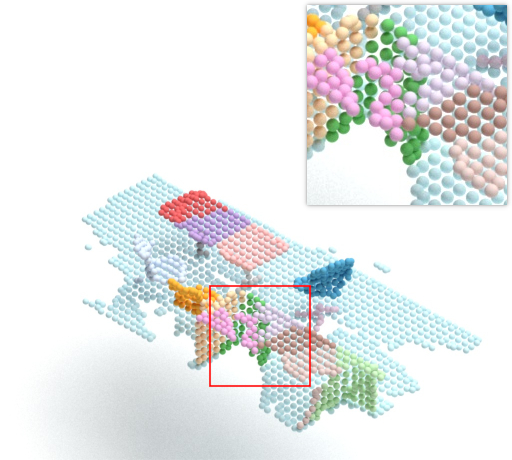}
    % \caption{Instance label}
    % \label{fig:boundary_structure}
  \end{subfigure}
  \hfill
  \begin{subfigure}[b]{0.18\linewidth}
    \centering
    \includegraphics[width=\textwidth]{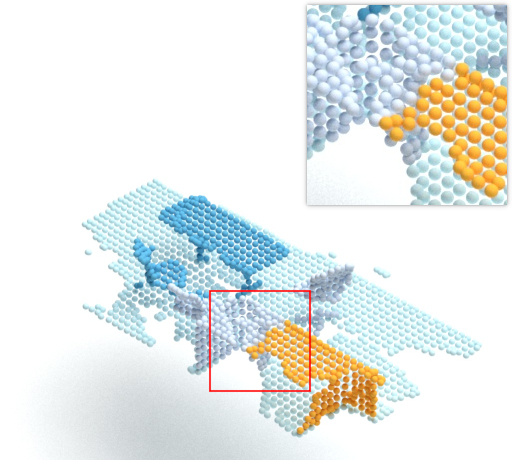}
    % \caption{Our segmentation}
    % \label{fig:boundary_structure}
  \end{subfigure}
  \begin{subfigure}[b]{0.18\linewidth}
    \centering
    \includegraphics[width=\textwidth]{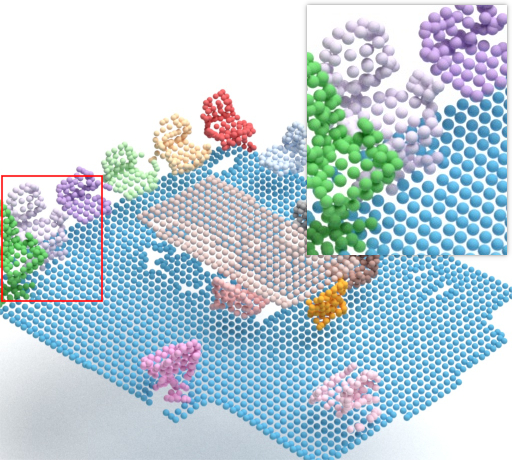}
    \caption{GT Label}
    % \label{fig:boundary_structure}
  \end{subfigure}
  \hfill
  \begin{subfigure}[b]{0.18\linewidth}
    \centering
    \includegraphics[width=\textwidth]{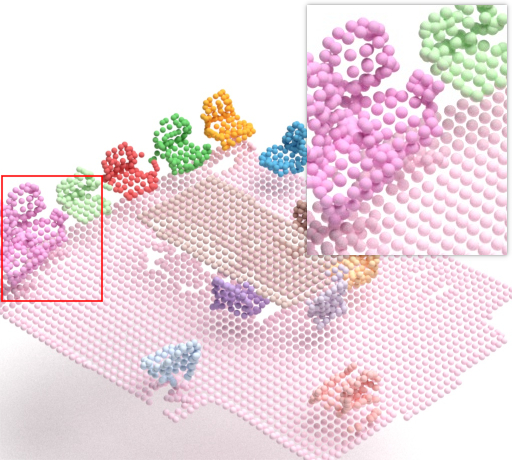}
    \caption{PG}
    % \label{fig:boundary_structure}
  \end{subfigure}
  \hfill
  \begin{subfigure}[b]{0.18\linewidth}
    \centering
    \includegraphics[width=\textwidth]{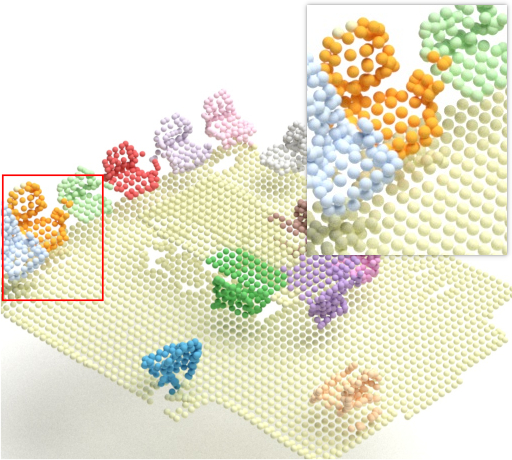}
    \caption{Ours}
    % \label{fig:boundary_structure}
  \end{subfigure}
  \hfill
  \begin{subfigure}[b]{0.18\linewidth}
    \centering
    \includegraphics[width=\textwidth]{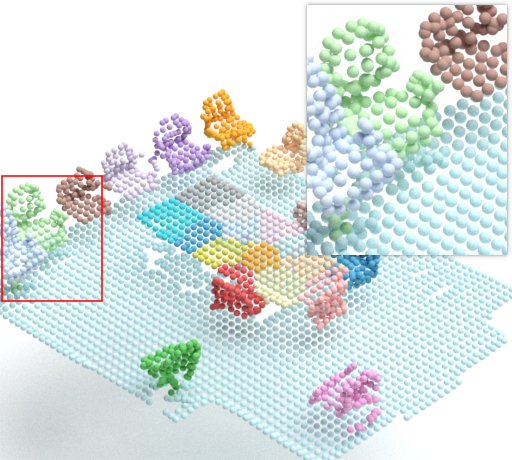}
    \caption{MS $0.06$}
    % \label{fig:boundary_structure}
  \end{subfigure}
  \hfill
  \begin{subfigure}[b]{0.18\linewidth}
    \centering
    \includegraphics[width=\textwidth]{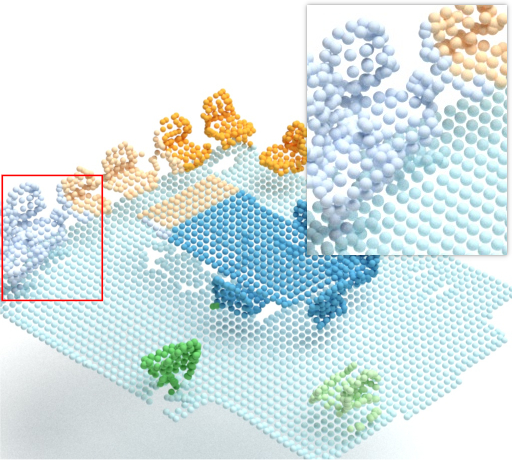}
    \caption{MS $0.15$}
    % \label{fig:boundary_structure}
  \end{subfigure}
  %%%%%%%%%%%%%%%%%%%%%%%%%%%%%%%%%%%%%%%%%%%%%%%%%%%%%%%%%%%%%%%%%%%%%%%%%%%%%%%%%
  \caption{S3DIS segmentation results.}
  \label{fig:S3DIS}
  \vspace{-5mm}
\end{figure*}

\noindent \textbf{Evaluation.}
Per-class mIoU are reported in Table \ref{table:s3dis_results}.
Since the number of objects of different types varies greatly, we thus additionally report macro-average to show the overall performance of different models.
Not surprisingly, Point Group (PG) achieves the highest mIoU across all categories.
Similar to our model, Point Group mis-classifies object points that are close to the floor as floor category (row 1 Fig. \ref{fig:S3DIS}).
Mean-shift (MS) does not share the same vulnerability since floors are manually detected and removed.
However, Mean-shift is sensitive to the bandwidth value.
The bandwidth reflects the prior object sizes.
With the bandwidth tuned for Chair (MS 0.06) whose sizes are small on average, tables and sofas are largely over-segmented. 
With the bandwidth tuned for Table (MS 0.15), chairs tend to be under-segmented.

As demonstrated in row 3 Fig. \ref{fig:S3DIS}, chairs are successfully segmented by our model even when multiple chairs are clustered together.
The segmentation of tables presents a challenge for our model since their sizes are commonly larger than our maximum glimpses sizes.
However, SPAIR3D still tries to expand the glimpses sizes to better model larger objects while keeping the total number of glimpses low (row 2 Fig. \ref{fig:S3DIS}).

The experimental results in Table~\ref{table:s3dis_results} demonstrate the potential of applying a generative model to more complicated scenarios. 

\begin{table}
  \begin{center}
  \begin{tabular}{c|c|c|c|c|c}
           &     & Chair $\uparrow$ & Table $\uparrow$  & Sofa  $\uparrow$  & macro-avg  $\uparrow$ \\ \cline{1-1} \hline
  PG       & S   & $0.61$           & $0.69$            & $0.52$            & $0.60$                \\ \hline
  MS 0.06  & U   & $0.75$           & $0.34$            & $0.36$            & $0.48$                \\ \hline
  MS 0.15  & U   & $0.33$           & $0.46$            & $0.38$            & $0.39$                \\ \hline
  SPAIR3D (ours)     & U   & $0.59$           & $0.43$            & $0.49$            & $0.50$                \\ \hline

  \end{tabular}    
  \vspace{2mm}    
  \caption{Segmentation results on S3DIS. 'S' and 'U' denote the corresponding models are trained in supervised and unsupervised manner, respectively. }
  \label{table:s3dis_results}
  \vspace{-8mm}
\end{center}
\end{table}

% \begin{table}
%   \begin{center}
%   \begin{tabular}{c|c|c|c|c}
%             & Chair $\uparrow$ & Table $\uparrow$  & Sofa  $\uparrow$  & macro-avg  $\uparrow$ \\ \cline{1-1} \hline
%   PG        & $0.61$           & $0.69$            & $0.52$            & $0.60$                \\ \hline
%   MS 0.06   & $0.75$           & $0.34$            & $0.36$            & $0.48$                \\ \hline
%   MS 0.15   & $0.33$           & $0.46$            & $0.38$            & $0.39$                \\ \hline
%   Ours      & $0.59$           & $0.43$            & $0.49$            & $0.50$                \\ \hline

%   \end{tabular}     
%   \vspace{3mm}   
%   \caption{3D point cloud segmentation results on S3DIS.}
%   \label{table:s3dis_results}
%   \vspace{-8mm}
% \end{center}
% \end{table}

\subsection{Ablation Study of Multi-layer PointGNN}
\begin{figure}
  \centering
  \begin{subfigure}[b]{0.18\linewidth}
      \centering
      \includegraphics[width=\textwidth]{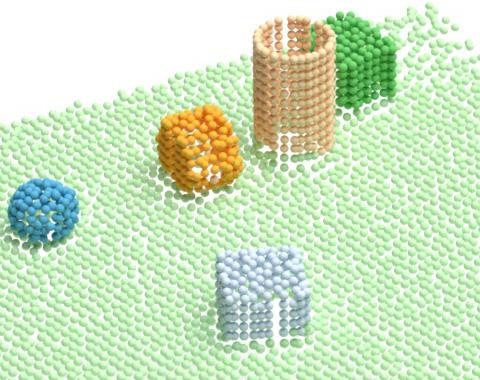}
      \caption{With PointGNNs}
  \end{subfigure}
  \hspace{0.5cm}
  \begin{subfigure}[b]{0.18\linewidth}
      \centering
      \includegraphics[width=\textwidth]{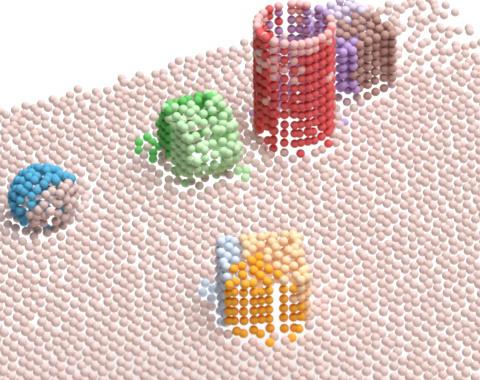}
      \caption{w/o PointGNNs}
  \end{subfigure}
  \caption{The comparison between models with (left) and without (right) multi-layer PointGNNs.~It shows that objects are over-segmented severely without multi-layer PointGNNs.}
  \label{fig:ZPres}
  \vspace{-5mm}
\end{figure}

To evaluate the importance of multi-layer PointGNN in $\mathbf{z}^{pres}_i$
generation (right branch in Fig.~\ref{fig:glimpsevae}), we remove the
multi-layer PointGNN and generate $\mathbf{z}^{pres}_i$ directly from
$\mathbf{f}_i$.
The ablated model on the UOR dataset achieves \textbf{ARI:}$\mathbf{0.841}$, \textbf{SC:}$\mathbf{0.610}$, and \textbf{mSC:}$\mathbf{0.627}$,
which is significantly worse than the full SPAIR3D model.
The performance distribution of ablated SPAIR3D (Fig~\ref{fig:test_distribution}, first row) indicates that removing the multi-layer PointGNN has a negative influence on the entire dataset.
Fig.~\ref{fig:ZPres} shows that the multi-layer PointGNN is crucial to preventing over-segmentation.

\subsection{Empirical Evaluation of PGD}
\begin{figure}[h]
  \begin{center}
     \begin{tabular}{cccccc}
        \includegraphics[width=0.13\linewidth]{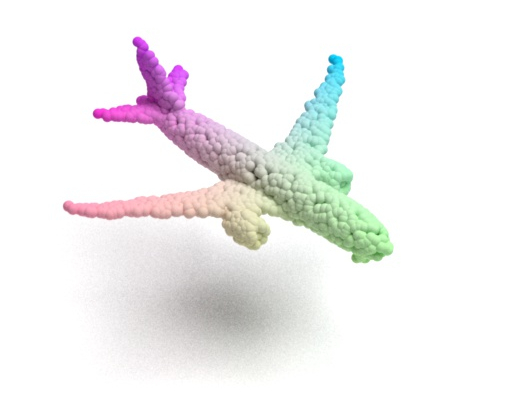}
        & \includegraphics[width=0.13\linewidth]{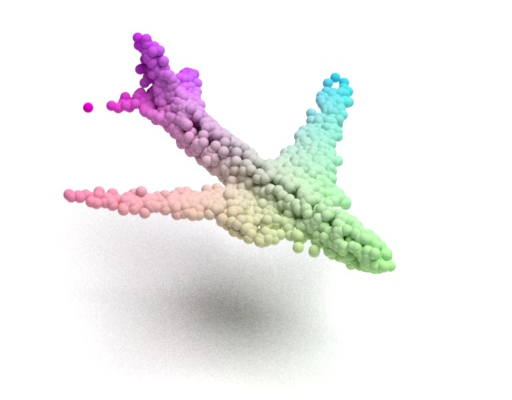}
        & \includegraphics[width=0.13\linewidth]{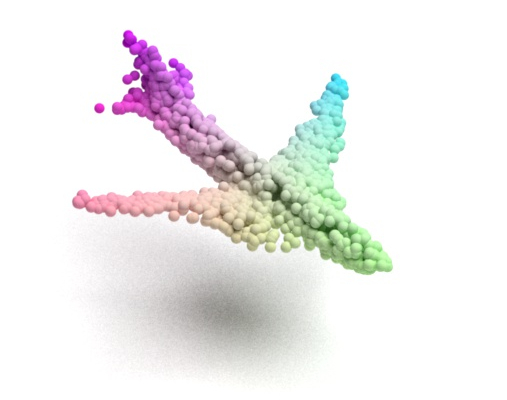}
        & \includegraphics[width=0.13\linewidth]{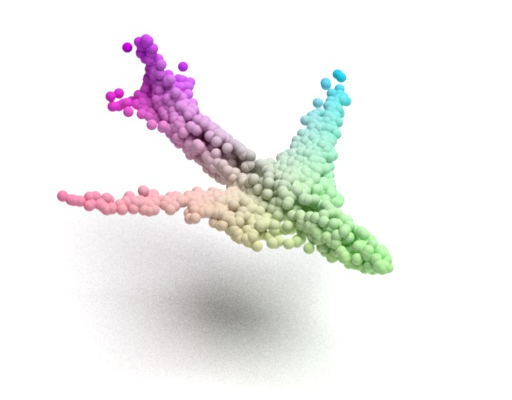}
        & \includegraphics[width=0.13\linewidth]{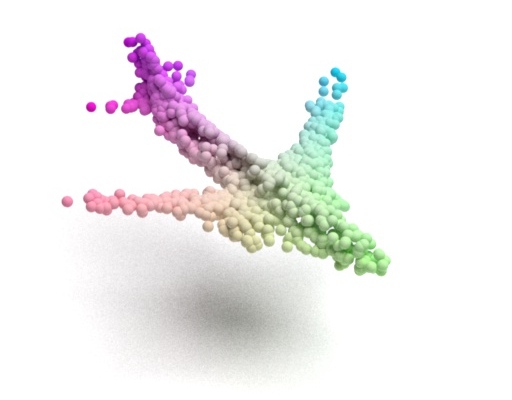}
        & \includegraphics[width=0.13\linewidth]{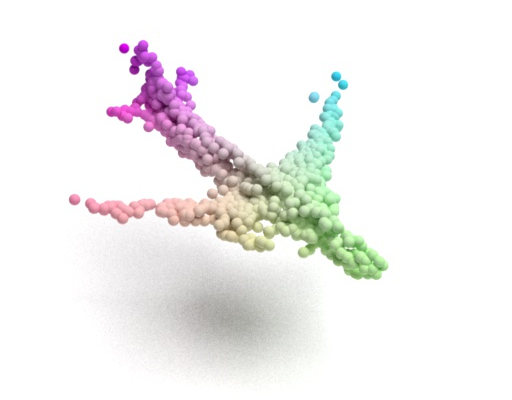}\\   
        (a)&(b)&(c)&(d)&(e)&(f)
     \end{tabular}
  \end{center}
  \caption{PGD trained on ShapeNet. (a) Input point cloud with $N$ points. Reconstruction with (b) $1.5N$, (c) $1.25N$, (d) $N$, (e) $0.75N$ , and (f) $0.5N$ points.}
  \label{fig:PGF}
  \vspace{-1cm}
\end{figure}

3D objects of the same category can be modeled by a varying number of points. 
The generation quality of the point cloud largely depends on the robustness of our model against the number of points representing each object.
To demonstrate that PGD can reconstruct each object with a dynamic number of points, we train the global VAE on the ShapeNet dataset~\cite{shapenet}, where each object is composed of roughly $2000$ points, and reconstruct the object with a varying number of points.
For reference input point clouds of size $N$, we force PGD to reconstruct a point cloud of size $1.5N$, $1.25N$, $N$, $0.75N$, and $0.5N$, respectively. 
As shown in Fig.~\ref{fig:PGF}, while with fewer details compared to the input, the reconstructions capture the overall object structure in all $5$ settings.

\section{Limitations}

Similar to SPAIR, the good scalability in SPAIR3D stems from the local attention and reconstruction mechanisms.
By design, each voxel cell can only propose one object.
Thus, it is difficult to detect multiple objects that exist in the same voxel cell.
If one object is much larger than the size of the voxel cells, no voxel cells can accurately infer complete object information from its local perceptive field. 
One can alleviate the problem with overlapping voxel cells and make the mixture model hierarchical, which we leave as future work.
% Given these limitations are inherent to the SPAIR framework, they can really only be resolved at a meta-level.
% In our intended use case, an intelligent agent that employs SPAIR3D as a subroutine can use voxel cells of vastly different sizes to segment out objects of different sizes and use a spatial knowledge representation and reasoning formalism [\cite{COHN08}] to resolve inconsistencies and get to a full scene representation. 

\section{Conclusion and Future Work}
Our proposed \emph{SPAIR3D} algorithm is, to the best of our knowledge, the first generative unsupervised object-centric learning model on point cloud with applications to 3D object segmentation tasks. 
The experimental results demonstrate that SPAIR3D can generalize well to previously unseen scenes with a large number of objects without performance degeneration.
The spatial mixture interpretation of SPAIR3D opens up the possibility to other extensions including memory mechanism \cite{VMA} or iterative refinement \cite{IODINE}, which is left as our future work.
% Our model may tend to over-segment objects of a significant different scale from those in the training dataset, which is left as our future work. 

\clearpage
\bibliographystyle{splncs04}
\bibliography{egbib}

\begin{thebibliography}{10}
\providecommand{\url}[1]{\texttt{#1}}
\providecommand{\urlprefix}{URL }
\providecommand{\doi}[1]{https://doi.org/#1}

\bibitem{achlioptas2018learning}
Achlioptas, P., Diamanti, O., Mitliagkas, I., Guibas, L.: Learning
  representations and generative models for 3d point clouds. In: ICML. pp.
  40--49. PMLR (2018)

\bibitem{DVIB}
Alemi, A., Fischer, I., Dillon, J., Murphy, K.: Deep variational information
  bottleneck. In: ICLR (2017), \url{https://arxiv.org/abs/1612.00410}

\bibitem{S3DIS}
{Armeni}, I., {Sax}, A., {Zamir}, A.R., {Savarese}, S.: {Joint 2D-3D-Semantic
  Data for Indoor Scene Understanding}. ArXiv e-prints  (Feb 2017)

\bibitem{VMA}
Bornschein, J., Mnih, A., Zoran, D., Rezende, D.J.: Variational memory
  addressing in generative models. In: NIPS (2017)

\bibitem{MONET}
Burgess, C., Matthey, L., Watters, N., Kabra, R., Higgins, I., Botvinick, M.,
  Lerchner, A.: Monet: Unsupervised scene decomposition and representation.
  ArXiv  \textbf{abs/1901.11390} (01 2019),
  \url{https://arxiv.org/abs/1901.11390}

\bibitem{UbetaVAE}
Burgess, C.P., Higgins, I., Pal, A., Matthey, L., Watters, N., Desjardins, G.,
  Lerchner, A.: Understanding disentangling in beta-vae. ArXiv
  \textbf{abs/1804.03599} (2018)

\bibitem{shapenet}
Chang, A.X., Funkhouser, T., Guibas, L., Hanrahan, P., Huang, Q., Li, Z.,
  Savarese, S., Savva, M., Song, S., Su, H., Xiao, J., Yi, L., Yu, F.:
  {ShapeNet: An Information-Rich 3D Model Repository}. Tech. Rep.
  arXiv:1512.03012 [cs.GR], Stanford University --- Princeton University ---
  Toyota Technological Institute at Chicago (2015)

\bibitem{OOPhysics1}
Chang, B.M., Ullman, T., Torralba, A., Tenenbaum, B.J.: A compositional
  object-based approach to learning physical dynamics. ICLR  (2017)

\bibitem{ROOTS}
Chen, C., Deng, F., Ahn, S.: Roots: Object-centric representation and rendering
  of 3d scenes (2021)

\bibitem{SPAIR}
Crawford, E., Pineau, J.: Spatially invariant unsupervised object detection
  with convolutional neural networks. AAAI  \textbf{33},  3412--3420 (07 2019).
  \doi{10.1609/aaai.v33i01.33013412}

\bibitem{SPAIROT}
Crawford, E., Pineau, J.: Exploiting spatial invariance for scalable
  unsupervised object tracking. AAAI  \textbf{34},  3684--3692 (04 2020).
  \doi{10.1609/aaai.v34i04.5777}

\bibitem{OOMDP}
Diuk, C., Cohen, A., Littman, M.L.: An object-oriented representation for
  efficient reinforcement learning. In: ICML. p. 240–247. ICML ’08,
  Association for Computing Machinery, New York, NY, USA (2008).
  \doi{10.1145/1390156.1390187}, \url{https://doi.org/10.1145/1390156.1390187}

\bibitem{GENESIS}
Engelcke, M., Kosiorek, A.R., Jones, O.P., Posner, I.: Genesis: Generative
  scene inference and sampling with object-centric latent representations. In:
  ICLR (2020), \url{https://openreview.net/forum?id=BkxfaTVFwH}

\bibitem{AIR}
Eslami, S.M.A., Heess, N., Weber, T., Tassa, Y., Szepesvari, D., Kavukcuoglu,
  K., Hinton, G.E.: Attend, infer, repeat: Fast scene understanding with
  generative models. In: Proceedings of the 30th International Conference on
  Neural Information Processing Systems. p. 3233–3241. NIPS'16, Curran
  Associates Inc., Red Hook, NY, USA (2016)

\bibitem{gadelha2018multiresolution}
Gadelha, M., Wang, R., Maji, S.: Multiresolution tree networks for 3d point
  cloud processing. In: ECCV. pp. 103--118 (2018)

\bibitem{IODINE}
Greff, K., Kaufman, R.L., Kabra, R., Watters, N., Burgess, C., Zoran, D.,
  Matthey, L., Botvinick, M.M., Lerchner, A.: Multi-object representation
  learning with iterative variational inference. In: ICML (2019)

\bibitem{NEM}
Greff, K., van Steenkiste, S., Schmidhuber, J.: Neural expectation
  maximization. In: NeurIPS. p. 6694–6704. NIPS'17, Curran Associates Inc.,
  Red Hook, NY, USA (2017)

\bibitem{ObjectCentricVideoGeneration}
Henderson, P., Lampert, C.H.: Unsupervised object-centric video generation and
  decomposition in {3D}. In: NeurIPS (2020)

\bibitem{betaVAE}
Higgins, I., Matthey, L., Pal, A., Burgess, C., Glorot, X., Botvinick, M.M.,
  Mohamed, S., Lerchner, A.: beta-vae: Learning basic visual concepts with a
  constrained variational framework. In: ICLR (2017)

\bibitem{ARI}
Hubert, L., Arabie, P.: Comparing partitions. Journal of Classification
  \textbf{2},  193--218 (1985)

\bibitem{CNN}
Islam*, M.A., Jia*, S., Bruce, N.D.B.: How much position information do
  convolutional neural networks encode? In: International Conference on
  Learning Representations (2020),
  \url{https://openreview.net/forum?id=rJeB36NKvB}

\bibitem{PointGroup}
Jiang, L., Zhao, H., Shi, S., Liu, S., Fu, C.W., Jia, J.: Pointgroup: Dual-set
  point grouping for 3d instance segmentation. CVPR  (2020)

\bibitem{CLEVR}
Johnson, J., Hariharan, B., van~der Maaten, L., Fei-Fei, L., Zitnick, C.,
  Girshick, R.: Clevr: A diagnostic dataset for compositional language and
  elementary visual reasoning. pp. 1988--1997 (07 2017).
  \doi{10.1109/CVPR.2017.215}

\bibitem{Unity}
Juliani, A., Berges, V., Teng, E., Cohen, A., Harper, J., Elion, C., Goy, C.,
  Gao, Y., Henry, H., Mattar, M., Lange, D.: Unity: A general platform for
  intelligent agents. ArXiv  \textbf{abs/1809.02627} (2020)

\bibitem{multiobjectdatasets19}
Kabra, R., Burgess, C., Matthey, L., Kaufman, R.L., Greff, K., Reynolds, M.,
  Lerchner, A.: Multi-object datasets.
  https://github.com/deepmind/multi-object-datasets/ (2019)

\bibitem{ObjectFile}
Kahneman, D., Treisman, A., Gibbs, B.J.: The reviewing of object files:
  Object-specific integration of information. Cognitive Psychology
  \textbf{24}(2),  175 -- 219 (1992).
  \doi{https://doi.org/10.1016/0010-0285(92)90007-O},
  \url{http://www.sciencedirect.com/science/article/pii/001002859290007O}

\bibitem{SchemaNetworks}
Kansky, K., Silver, T., M{\'e}ly, D.A., Eldawy, M., L{\'a}zaro-Gredilla, M.,
  Lou, X., Dorfman, N., Sidor, S., Phoenix, D.S., George, D.: Schema networks:
  Zero-shot transfer with a generative causal model of intuitive physics. ArXiv
   \textbf{abs/1706.04317} (2017)

\bibitem{VAE}
Kingma, D.P., Welling, M.: {Auto-Encoding Variational Bayes}. In: 2nd
  International Conference on Learning Representations, {ICLR} 2014, Banff, AB,
  Canada, April 14-16, 2014, Conference Track Proceedings (2014)

\bibitem{MulMON}
Li, N., Eastwood, C., Fisher, R.: Learning object-centric representations of
  multi-object scenes from multiple views. In: Larochelle, H., Ranzato, M.,
  Hadsell, R., Balcan, M.F., Lin, H. (eds.) Advances in Neural Information
  Processing Systems. vol.~33, pp. 5656--5666. Curran Associates, Inc. (2020),
  \url{https://proceedings.neurips.cc/paper/2020/file/3d9dabe52805a1ea21864b09f3397593-Paper.pdf}

\bibitem{SPACE}
Lin, Z., Wu, Y.F., Peri, S.V., Sun, W., Singh, G., Deng, F., Jiang, J., Ahn,
  S.: Space: Unsupervised object-oriented scene representation via spatial
  attention and decomposition. In: ICLR (2020),
  \url{https://openreview.net/forum?id=rkl03ySYDH}

\bibitem{SlotAttention}
Locatello, F., Weissenborn, D., Unterthiner, T., Mahendran, A., Heigold, G.,
  Uszkoreit, J., Dosovitskiy, A., Kipf, T.: Object-centric learning with slot
  attention. In: NeurIPS (2020)

\bibitem{PointMDP}
Luo, S., Hu, W.: Diffusion probabilistic models for 3d point cloud generation.
  In: Proceedings of the IEEE/CVF Conference on Computer Vision and Pattern
  Recognition (CVPR) (June 2021)

\bibitem{tSNE}
van~der Maaten, L., Hinton, G.: Visualizing data using {t-SNE}. Journal of
  Machine Learning Research  \textbf{9},  2579--2605 (2008),
  \url{http://www.jmlr.org/papers/v9/vandermaaten08a.html}

\bibitem{PointGNN}
{Shi}, W., {Rajkumar}, R.: Point-gnn: Graph neural network for 3d object
  detection in a point cloud. In: CVPR. pp. 1708--1716 (2020)

\bibitem{ObSuRF}
Stelzner, K., Kersting, K., Kosiorek, A.R.: Decomposing 3d scenes into objects
  via unsupervised volume segmentation (2021)

\bibitem{IB}
Tishby, N., Pereira, F.C., Bialek, W.: The information bottleneck method. In:
  Proc. of the 37-th Annual Allerton Conference on Communication, Control and
  Computing. pp. 368--377 (1999), \url{https://arxiv.org/abs/physics/0004057}

\bibitem{PointConv}
Wu, W., Qi, Z., Li, F.: Pointconv: Deep convolutional networks on 3d point
  clouds. In: CVPR. pp. 9613--9622 (06 2019). \doi{10.1109/CVPR.2019.00985}

\bibitem{PointFlow}
Yang, G., Huang, X., Hao, Z., Liu, M.Y., Belongie, S., Hariharan, B.:
  Pointflow: 3d point cloud generation with continuous normalizing flows. pp.
  4540--4549 (10 2019). \doi{10.1109/ICCV.2019.00464}

\end{thebibliography}

% \appendix

% \section{123}

\end{document}